\pdfoutput=1
\documentclass{article}
\usepackage{iclr2027_conference,times}

\usepackage{amsmath,amsfonts,bm}

\def\eqref#1{equation~\ref{#1}}

\def\1{\bm{1}}

\def\eps{{\epsilon}}

\DeclareMathAlphabet{\mathsfit}{\encodingdefault}{\sfdefault}{m}{sl}
\SetMathAlphabet{\mathsfit}{bold}{\encodingdefault}{\sfdefault}{bx}{n}

\newcommand{\E}{\mathbb{E}}

\DeclareMathOperator*{\argmin}{arg\,min}

\usepackage[table]{xcolor}
\usepackage{hyperref}
\usepackage{url}
\usepackage{amsmath,amssymb,amsthm}
\usepackage{needspace}
\usepackage{algorithm}
\usepackage{algpseudocode}
\usepackage{booktabs}
\usepackage{longtable}
\usepackage{arydshln}
\usepackage{multirow}
\usepackage{pifont}
\usepackage{graphicx}
\usepackage{enumitem}
\usepackage{placeins}
\usepackage{tcolorbox}
\usepackage{fvextra}
\usepackage{tikz}
\usetikzlibrary{arrows.meta,positioning}
\usepackage{wrapfig}
\usepackage{capt-of}
\usepackage{etoc}
\newtcolorbox{artifactbox}[1]{
  colback=blue!3!white, colframe=blue!30!black!60,
  boxrule=0.5pt, arc=2.5pt,
  left=7pt, right=7pt, top=5pt, bottom=5pt,
  colbacktitle=blue!8!white, coltitle=black,
  fonttitle=\bfseries\small, title={#1}}
\tcbuselibrary{breakable,listings,skins}
\DeclareTextSymbolDefault{\textquotedbl}{T1}%
\DeclareTextSymbolDefault{\textasciitilde}{T1}
\lstdefinestyle{artifactcode}{upquote=true, literate={~}{{\textasciitilde}}1, language=Python, basicstyle=\ttfamily\scriptsize, keywordstyle=\color{blue!55!black}\bfseries, stringstyle=\color{green!35!black}, commentstyle=\color{black!50}\itshape, columns=fixed, keepspaces=true, breaklines=true, breakatwhitespace=false, showstringspaces=false, tabsize=4, postbreak=\mbox{$\hookrightarrow$\space}}
\lstdefinestyle{artifacttext}{upquote=true, literate={~}{{\textasciitilde}}1, basicstyle=\ttfamily\scriptsize, columns=fullflexible, keepspaces=true, breaklines=true, showstringspaces=false, postbreak=\mbox{$\hookrightarrow$\space}}
\tcbset{artifactstyle/.style={breakable, enhanced jigsaw, listing only, colback=blue!3!white, colframe=blue!30!black!60, boxrule=0.5pt, arc=2.5pt, left=6pt, right=6pt, top=3pt, bottom=3pt, colbacktitle=blue!8!white, coltitle=black, fonttitle=\bfseries\small, before skip=6pt, after skip=8pt}}
\newtcblisting{codeblock}[1]{artifactstyle, title={#1}, listing options={style=artifactcode}}
\newtcblisting{textblock}[1]{artifactstyle, title={#1}, listing options={style=artifacttext}}
\newcommand{\method}{VERSE}
\newcommand{\Str}{S_{\mathrm{train}}}
\newcommand{\Sval}{S_{\mathrm{val}}}
\newcommand{\Stest}{S_{\mathrm{test}}}
\newcommand{\HH}{\mathcal H}
\newcommand{\Prb}{\mathbb P}
\newcommand{\CC}{\mathcal C}
\newcommand{\Aff}{\operatorname{Aff}}
\renewcommand{\eps}{\varepsilon}
\theoremstyle{plain}
\newtheorem{theorem}{Theorem}
\newtheorem{lemma}{Lemma}[subsection]
\theoremstyle{definition}
\newtheorem{proposition}[lemma]{Proposition}
\newtheorem*{remark}{Remark}
\newenvironment{formaltheorem}[1]
  {\par\addvspace{\topsep}\noindent\textbf{Theorem #1 (formal).}\ \itshape}
  {\par\addvspace{\topsep}}

\newcommand{\yes}{\checkmark}
\newcommand{\no}{\ding{55}}

\title{{VERSE}: Verified Self-Evolving Optimizer for Agent Harnesses}

\author{
\stepcounter{footnote}
\hspace{-5.5pt}\textbf{Zekai Wang$^{1,}$\thanks{Work done during an internship at Amazon.\newline\hspace*{1.8em}Correspondence to: Zekai Wang \textless zekai@mit.edu\textgreater, Yingqiang Ge \textless gyq@amazon.com\textgreater.}$\;\,$, Yingqiang Ge$^{2}$, Zekun Wang$^{2,}$\footnotemark[\value{footnote}]$\;\,$, Hai Wang$^{2}$, Yuhui Xu$^{2}$,} \vspace{0.05in}\\
\textbf{Joshua Frandsen$^{2}$, Shancong Fu$^{2}$, Ashia C. Wilson$^{1}$, Chandan K. Reddy$^{2}$}\vspace{0.05in} \\ 
 \ $^{1}$MIT \hspace{2.5pt} $^{2}$Amazon\vspace{-0.1in}
}

\iclrfinalcopy
\makeatletter
\newcommand{\LTnormalcaption}{%
  \setlength{\LTcapwidth}{\textwidth}%
  \let\LT@makecaption@orig\LT@makecaption
  \def\LT@makecaption##1##2##3{\LT@makecaption@orig{##1}{\normalsize##2}{\normalsize##3}}}
\makeatother
\begin{document}

\maketitle
\lhead{Preprint}

\begin{abstract}
Harness evolution improves an LLM agent's prompts, tools, and workflow, while the optimizer's own tools and procedures often remain fixed. We study whether an optimizer can improve another agent more effectively by also improving how it diagnoses failures, develops edits, and tests their effects. Two observations guide our design. In a controlled study, optimizer self-evolution fails to improve performance without execution-based verification, but achieves the best result of that study when verification is available. Across five executors, self-evolving optimizers build their own tools for failure analysis, verification, training audits, and workflow control. Motivated by these findings, we introduce \method{}, a \textbf{Ver}ified \textbf{S}elf-\textbf{E}volving optimizer for agent harnesses. \method{} lets the optimizer test draft edits, replay failures, and perturb suspected steps before submission, while tracking fixes and regressions across rounds. Using this feedback, the optimizer revises both the executor harness and its own prompts, skills, tools, hooks, and notes, while the weights of the optimizer and executor models stay fixed. Under a shared protocol with disjoint training, validation, and test tasks, \method{} improves all four evaluated harness optimizers on held-out SWE-rebench tasks and newer out-of-distribution tasks in five languages. Its best validation-selected harness reaches $42.3\%$ and $37.7\%$ accuracy, respectively, against $39.2\%$ and $29.3\%$ for the strongest baselines. Code is available at \url{https://github.com/wzekai/VERSE}.
\end{abstract}

\etocdepthtag.toc{mtchapter}
\section{Introduction}
\label{sec:intro}

An LLM agent is a model wrapped in a harness: the prompts, tools, memory, and control flow that turn a language model into a system that acts. On agentic benchmarks such as SWE-bench~\citep{jimenez2024swebench} and Terminal-Bench~\citep{terminalbench2026}, the harness often matters as much as the model. In \emph{harness evolution}, an LLM optimizer reads the failed trajectories of a frozen executor agent and edits the executor's harness, over several rounds~\citep{metaharness2026,ahe2026,selfharness2026,harnessx2026,tthe2026}. The optimizer's own harness, however, stays fixed: it is a hand-built pipeline or an off-the-shelf coding agent~\citep{metaharness2026,ahe2026,harnessx2026,autosaddler2026}, or, when one model plays both roles, a fixed loop that proposes and accepts edits~\citep{selfharness2026}. Lessons from earlier rounds shape the executor's next harness, but not how the optimizer works. Can the optimizer evolve its own harness as well?

Simply allowing it did not help (Section~\ref{sec:observations}). When we added self-evolution to Meta-Harness~\citep{metaharness2026}, the optimizer wrote untested lessons into its own harness, and no later round beat the initial harness on validation. The optimizer learns only from trajectories and scores, and it sees the effect of an edit one round later. Reading trajectories can mislead: optimizers can report failures that never occurred~\citep{phantom2026}, and evolved harnesses can gain little on new tasks~\citep{rethinking2026}. Once the same optimizer could run a draft edit on a training task before submitting it, self-evolution gave the best result among the settings we compared. Execution checks give a self-evolving optimizer evidence about which of its lessons hold.

A second study shows what a self-evolving optimizer builds for itself. Across five executors, it tends to build tools that find the causes of failures, checks that a draft works, notes on which fixes held, and rules for its own workflow. Some recent self-improving systems also revise their own improvement procedure~\citep{wang2026selfconsolidating,zhang2026hyperagents,zhou2026hsi} (Appendix~\ref{app:related}). Together, the two studies suggest a design: give the optimizer reliable execution checks from the first round, and let it evolve how it uses them.

These observations motivate our method, \method{}, which combines execution checks with optimizer self-evolution. \method{} provides from the first round what the self-evolving optimizer built for itself. Trace minimization cuts a failed trajectory down to the few steps that still reproduce the failure. We design three verification tools, and the optimizer decides when and how to call them: they run a draft on the tasks it targets, replay a recorded failure to check that it reproduces, and change a single step to test whether that step causes the failure. A training audit records which fixes held in later rounds and which broke. Between rounds, the optimizer revises its own prompts, skills, tools, hooks, and notes. This makes harness evolution a bilevel process: the inner loop edits the executor's harness, and the outer loop edits the optimizer's own harness (Section~\ref{sec:overview}).

\begin{figure}[t]
\centering
\includegraphics[width=\textwidth]{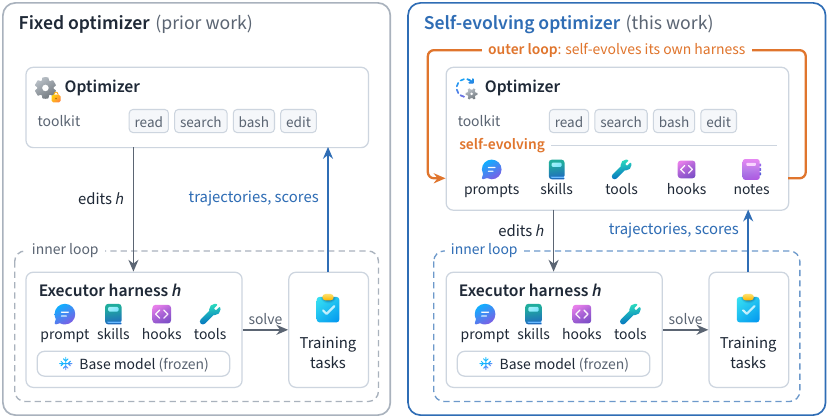}
\par\vspace{-4pt}
\caption{\textbf{Fixed versus self-evolving optimizers.} Left: in prior harness optimizers, the optimizer is fixed: it reads trajectories and scores, edits the executor harness $h$ that wraps a frozen base model, and keeps its own harness unchanged. Right: a self-evolving optimizer adds an outer loop: after each round, it updates its own prompts, skills, tools, hooks, and notes from the round's results.}
\label{fig:observation1}
\vspace{-6pt}
\end{figure}

We study this design in theory and in experiments. Our contributions are as follows.
\begin{itemize}[leftmargin=1.4em,itemsep=2pt,topsep=2pt]
\item Two studies: optimizer self-evolution helped only with execution-based verification, and the optimizer builds its own tools for failure analysis and verification. They motivate \method{}, which provides such tools and lets the optimizer evolve its own harness (Sections~\ref{sec:observations}--\ref{sec:method}).
\item A theoretical framework that views harness evolution as bilevel learning to generalize to new tasks. We prove that when the recorded evidence is ambiguous, reading trajectories alone leaves a minimum rate of wrong repairs, which executed checks can lower, and that wrong repairs limit what self-evolution can gain (Section~\ref{sec:theory}).
\item Experiments on SWE-rebench, where \method{} improves all four published harness optimizers it is added to, both on the test set and on newer out-of-distribution tasks. In the evolution logs, few draft edits fix their targets, and self-evolving optimizers add regression checks (Section~\ref{sec:experiments}).
\end{itemize}

\section{Setting and Motivating Observations}
\label{sec:observations}

\begin{figure}[t]
\centering
\includegraphics[width=\textwidth]{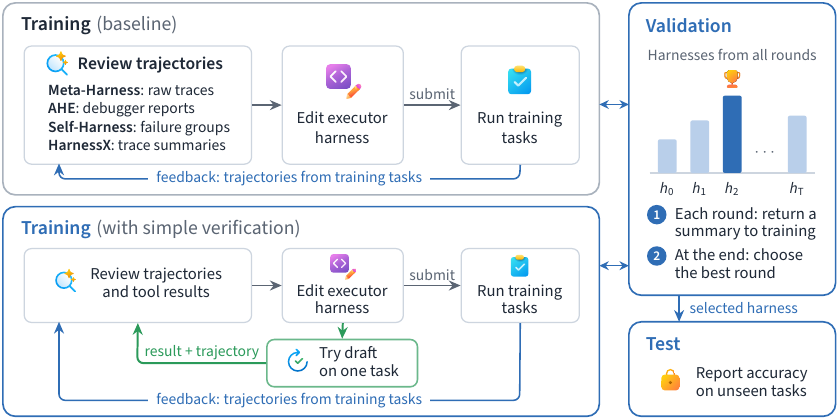}
\par\vspace{-4pt}
\caption{\textbf{Verification during harness evolution.} Prior optimizers review training trajectories differently (Appendix~\ref{app:baselines}), then edit the executor harness and rerun training tasks to collect new trajectories. Simple verification lets the optimizer execute a draft while editing and send its result and trajectory back for review, before submission and within budget. The validation summary has no task content, to limit overfitting (Section~\ref{sec:splits}). For the self-evolving optimizer (ours), an outer loop updates the optimizer's own harness between rounds (Figure~\ref{fig:observation1}).}
\label{fig:case}
\vspace{-8pt}
\end{figure}

We first describe the setup and the task splits used throughout the paper. We then present two studies that motivate \method{}. The first evaluates the role of verification in optimizer self-evolution. The second examines the tools, instructions, and notes that the optimizer builds for itself.

\subsection{Setup}
\label{sec:obs-setup}
A self-evolving optimizer can rewrite five parts of its own harness. \emph{Prompts}: the optimizer's system prompts. \emph{Skills}: instruction files added to the optimizer's prompt every round. \emph{Tools}: functions it writes and can call. \emph{Hooks}: code that runs automatically at fixed points of the optimizer's own loop, for example to check a tool call before it runs or to add a reminder after each turn. \emph{Notes}: its memory of earlier rounds. Self-written code is loaded only after it passes a safety check and a test run. Everything else is fixed: how harnesses are evaluated and selected, and the turn and execution budgets of each round. Keeping these fixed makes methods comparable under the same protocol.

For Observation 1, the optimizer is Qwen3.8-Flash-Next (176B total parameters, 6B active) and the executor is Qwen3.8-27B. Both stay frozen through six evolution rounds on SWE-rebench~\citep{swerebench2025}. We start from Meta-Harness~\citep{metaharness2026} and add simple verification (Figure~\ref{fig:case}), self-evolution (Figure~\ref{fig:observation1}), or both (Table~\ref{tab:observations}). For Observation 2, we run the self-evolving optimizer with simple verification with five executors and audit what it builds for itself.

\subsection{Training, validation, and test splits}
\label{sec:splits}
Many harness optimizers search and report results on the same benchmark~\citep{metaharness2026,ahe2026,selfharness2026,harnessx2026,zhang2026aevo}. \citet{rethinking2026} point out that such a search resembles test-time scaling, and find that its gains can come from overfitting to task-specific patterns rather than from a better harness design. We study how to evolve harnesses that generalize to unseen tasks. We therefore use three disjoint task sets, as is standard in machine learning~\citep{hastie2009elements}, and measure final performance only on tasks never used during evolution.

In the main text, we use SWE-rebench~\citep{swerebench2025}, a continuously refreshed pool of GitHub issue tasks in the SWE-bench format~\citep{jimenez2024swebench}. We draw 110 training and 50 validation tasks from a pool of 593 tasks in its monthly releases (January 2025 to February 2026). The test set is the 108 gradable tasks of the March 2026 release. All three sets consist of Python repositories, so we treat this test set as in-distribution. For an out-of-distribution (OOD) test set, we use the July 2026 release: its 107 gradable tasks are newer than every task in the three sets and span Go, Java, Python, Rust, and TypeScript, with only 20 in Python.

\begin{table}[t]
\caption{\textbf{Self-evolution with and without a verification tool} (SWE-rebench avg@3 accuracy, \%, $\pm$SEM over three test evaluations; one evolution run per row). \emph{Val-selected}: the harness selected by validation accuracy, from round $r^{\star}$; \emph{Last}: the final-round harness. At $r^{\star}=0$, validation selects $h_0$ itself, and the entry is an independent re-evaluation of it.}
\label{tab:observations}
\vspace{7pt}
\centering
\small
\setlength{\tabcolsep}{4pt}
\renewcommand{\arraystretch}{1.08}
\begin{tabular}{l cc c cc}
\toprule
Optimizer & Self-evolving & Verified & $r^{\star}$ & Val-selected & Last \\
\midrule
Blank (initial harness $h_0$) & -- & -- & -- & $33.95_{\pm0.82}$ & $33.95_{\pm0.82}$ \\
Meta-Harness~\citep{metaharness2026} & \no & \no & 4 & $39.20_{\pm1.11}$ & $35.80_{\pm0.82}$ \\
Self-evolving optimizer (w/o verification) & \yes & \no & 0 & $31.79_{\pm1.35}$ & $31.79_{\pm3.13}$ \\
Simple verification only & \no & \yes & 5 & $39.81_{\pm1.41}$ & $37.65_{\pm2.74}$ \\
\textbf{Self-evolving optimizer $+$ simple verification} & \yes & \yes & 6 & $\mathbf{41.05}_{\pm0.82}$ & $\mathbf{41.05}_{\pm0.82}$ \\
\bottomrule
\end{tabular}
\vspace{-10pt}
\end{table}

We select harnesses by validation accuracy, the percentage of validation tasks a harness resolves. This harness selection mirrors model selection in machine learning, where the checkpoint with the best validation performance is kept rather than the last one~\citep{prechelt1998early}. We report the validation-selected harness as our main result, together with the final-round harness. On the in-distribution test sets of Tables~\ref{tab:observations}--\ref{tab:ablation-open}, the validation-selected harness matches or beats the final-round harness in 15 of 18 runs, by up to 6.8 points, and trails it by at most 1.6 points. These results support validation-based selection over always reporting the final round.

Validation accuracy can also guide hyperparameter and architecture search~\citep{snoek2012practical,zoph2017neural}; both uses reduce the validation set to a single number. Our optimizer is a language model and, like other LLM-based optimizers~\citep{yuksekgonul2024textgrad,agrawal2026gepa}, can also learn from text feedback. After each round, we therefore give it a short text summary of its validation results: its validation accuracy in every round, the identifiers of the validation tasks its latest edit fixed or broke and whether each change held on a rerun, and how often its harness code raised errors. The optimizer never sees the content of validation tasks or their trajectories. This summary shows how well its edits generalize, which training accuracy cannot, because the optimizer has read every training trajectory. Validation therefore both guides later edits and selects the final harness, so validation accuracy can overstate test accuracy; Appendix~\ref{app:selection} bounds this gap by how much the feedback reveals about the validation set~\citep{dwork2015generalization,blum2015ladder}.

\subsection{Two Observations on Self-Evolving Optimizers}
\label{sec:two-observations}
\textbf{Observation 1: self-evolution helped with verification.}
Here, verification means letting the optimizer try an executor harness edit on a task \emph{before submitting it}. The baseline reads past trajectories, submits a new harness, and learns how it performs from the next evaluation. With simple verification, the optimizer can first run one training task with its edited harness, see whether the tests pass, and read the executor's trajectory. It can then change the harness and try again within the remaining call budget before submitting (Figure~\ref{fig:case}).

Without verification, adding self-evolution to Meta-Harness~\citep{metaharness2026} made it worse: no round beat round 0 on validation, so validation selected the initial harness over every evolved one ($r^{\star}=0$ in Table~\ref{tab:observations}). The optimizer could not test its edits before submitting them, so it could not tell which ones were wrong (Appendix~\ref{app:case} shows the failed edits and why they failed). This suggests that, without verification, a self-evolving optimizer can be misled by noisy validation results and write wrong lessons into its own harness. With the verification tool, the same optimizer could run a draft on a training task before submission and reached $41.05\%$ test accuracy, the highest in Table~\ref{tab:observations}.

\textbf{Observation 2: what the optimizer builds for itself.}
We run the self-evolving optimizer with simple verification (Figure~\ref{fig:case}) on SWE-rebench with five different executor models, and audit how the optimizer evolves its own harness to improve the executor's performance (Figure~\ref{fig:census}). Every run starts from an empty workspace. Appendix~\ref{app:artifacts} lists and classifies every artifact.

The five runs evolve 51 artifacts in total, each a prompt, skill, tool, hook, or notes file. Regardless of the executor, we find that the optimizer tends to build four kinds of artifacts. \emph{Attribution}: tools and skills that find and rank the causes of failures. \emph{Verification}: checks that a draft harness runs correctly and actually improves the executor. \emph{Training audit}: notes that track across rounds which tasks were fixed, which regressed, and which fixes were later undone. \emph{Workflow}: rules and hooks that change how the optimizer itself works, such as budget reminders and instructions for its parallel candidates.

\section{\method{}}
\label{sec:method}

Section~\ref{sec:observations} shows that self-evolution helped when the optimizer could verify its edits by execution, and that the optimizer repeatedly built its own tools for attribution, verification, and training audit. \method{} therefore provides these tools from the start and lets the optimizer adapt how it uses them.

\subsection{Overview of VERSE}
\label{sec:overview}
\label{sec:instruments}
\label{sec:self-evolve}
\label{sec:design-evidence}

\method{} is added to an existing harness optimizer, which we call the host, such as Meta-Harness~\citep{metaharness2026}. The hosts differ mainly in how they analyze failed trajectories and report the causes to the optimizer (Appendix~\ref{app:baselines}). \method{} keeps this report and adds four components on top: attribution, verification, a training audit, and self-evolution of the optimizer harness $g$ (Figure~\ref{fig:method}; pseudocode in Appendix~\ref{app:algorithm}).

\textbf{Attribution.}
Attribution proposes which steps of a failed training trajectory caused the failure. Failed trajectories are long, which makes the cause hard to find, so \method{} adds trace minimization: it re-runs subsets of its steps and keeps the few steps that still produce the same error. In our replay study, the median failed trajectory shrinks from 129 steps to 8 (Appendix~\ref{app:replay}).

\begin{figure}[t]
\centering
\includegraphics[width=\textwidth]{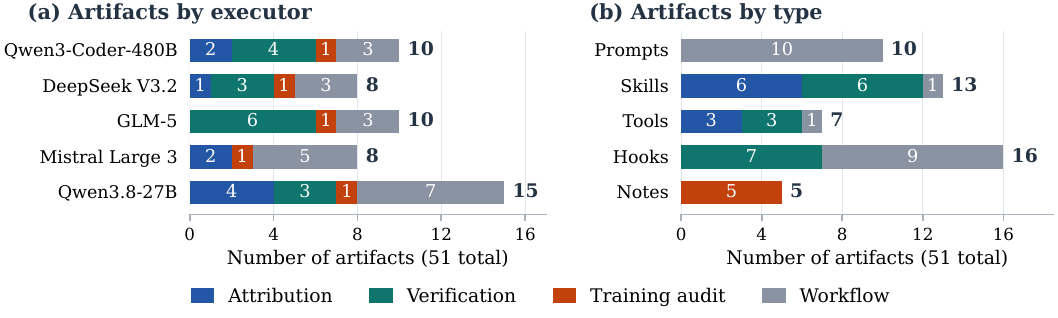}
\par\vspace{-10pt}
\caption{\textbf{What the optimizer builds for itself.} The self-evolving optimizer with simple verification runs six rounds on SWE-rebench with five open-weight executors (Appendix~\ref{app:artifacts} gives the setup of each run). Across these runs the optimizer evolved 51 artifacts for its own harness, which we audit: (a) artifacts by executor, colored by function; (b) the same artifacts by type.}
\label{fig:census}
\end{figure}

\textbf{Verification.}
Simple verification (Section~\ref{sec:observations}) runs a draft on one training task and leaves the result for the optimizer to interpret. \method{} adds three tools, each answering one question before submission. The \texttt{verification} tool asks whether a draft fixes the tasks it targets: it runs the draft on up to three of them and reports the failing tests and any errors in harness code. A task counts as fixed only if it passes twice, since a single pass can be luck. The \texttt{replay} tool asks whether a recorded failure is reproducible: it re-runs the recorded trajectory. The \texttt{perturbation} tool asks whether the failure depends on a given step: it removes or replaces that step in a reproducible trajectory and replays the rest. The first tool checks a fix; the other two check a diagnosis.

\textbf{Training audit.}
In Observation 2, every run kept round-by-round notes of its outcomes. \method{} keeps this record from the start. For each failure mode found by attribution, it tracks which training tasks still fail with it after every round. The optimizer can thus see which failures earlier edits removed and which returned.

\begin{figure}[t]
\centering
\includegraphics[width=\textwidth]{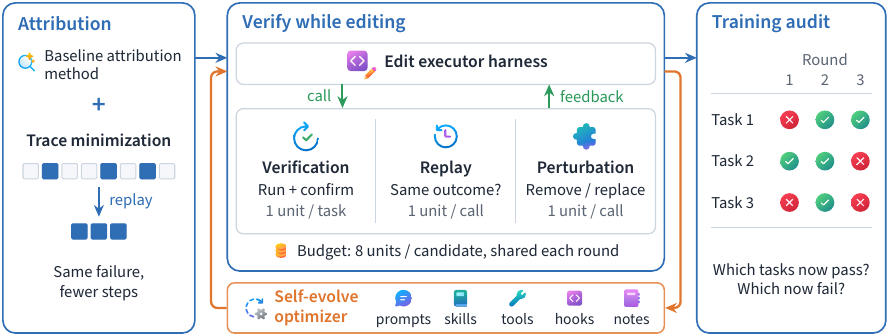}
\par\vspace{-4pt}
\caption{\textbf{Overview of \method{}.} Each round, attribution and the training audit report on the latest training runs, and the optimizer calls the verification tools while it edits the executor harness. Between rounds, the optimizer revises its own harness; the verification tools stay fixed, and it can build new tools on top of them.}
\label{fig:method}
\vspace{-4pt}
\end{figure}

\textbf{Optimizer self-evolution.}
At the end of each round, the optimizer spends up to 30 turns on a self-update. It reviews the round's results, including validation accuracy, whether the tasks it expected to fix were fixed, and its tool usage and errors. It then edits its own harness $g$ (prompts, skills, tools, hooks, and notes; Section~\ref{sec:obs-setup}). It also keeps simple verification, which shares the budget of the verification tools. These tools stay fixed, but its own code can call them.

\subsection{Theoretical Analysis}
\label{sec:theory}
\label{sec:formulation}
Improving a harness requires deciding what to fix and how to fix it. We analyze how verification helps identify the right repair, and how self-evolution can improve the search for a harness that performs well on new tasks. Formal statements and proofs appear in Appendix~\ref{app:theory}.

\textbf{Framework.}
A task $x$ includes a problem, an environment, and a success criterion. The executor has fixed model weights $\theta$ and a harness $h$ specifying its prompts, tools, and workflow. Its actions and the environment responses form a trajectory $y\sim\pi_\theta(\cdot\mid x;h)$. The loss $\ell(x,y)$ is $0$ for success and $1$ for failure. The population loss $L_{\theta,D}(h)$ averages this loss over new tasks $x\sim D$ and their execution outcomes. The optimizer has fixed weights $\omega$ and a separate harness $g$ consisting of its prompts, skills, tools, hooks, and notes (Section~\ref{sec:observations}); let $\mathcal T(g)$ denote its tools. The initial optimizer harness $g_0$ has tools $\mathcal T_0=\mathcal T(g_0)=\mathcal T_{\mathrm{basic}}\cup\mathcal T_{\mathrm{verify}}$, where
\begin{equation}
\begin{aligned}
\mathcal T_{\mathrm{basic}}&=\{\texttt{read},\ \texttt{search},\ \texttt{bash},\ \texttt{write\_notes},\ \texttt{submit\_proposal}\},\\
\mathcal T_{\mathrm{verify}}&=\{\texttt{verification},\ \texttt{replay},\ \texttt{perturbation}\}.
\end{aligned}
\label{eq:toolkit-theory}
\end{equation}
All methods share $\mathcal T_{\mathrm{basic}}$ (Appendix~\ref{app:theory}), and Section~\ref{sec:overview} describes $\mathcal T_{\mathrm{verify}}$. For a self-evolving optimizer, $\mathcal T_{\mathrm{verify}}$ also includes simple verification. Later optimizer harnesses keep all tools in $\mathcal T_0$, so $\mathcal T_0\subseteq\mathcal T(g)$, and can add new tools built from them.

We separate tasks into training, validation, and test sets $\Str,\Sval,\Stest$, with empirical losses $\widehat L_{\mathrm{train}},\widehat L_{\mathrm{val}},\widehat L_{\mathrm{test}}$ on these sets. In the inner loop, the optimizer edits and checks $h_t$ on training tasks, generating candidates $\CC_t(g)$ within budget $B_t$. Validation losses and outcome summaries from earlier rounds are available in both loops, for every method. In the outer loop, they and the round audits guide updates to $g$. The objectives are
\begin{equation}
\begin{aligned}
\text{inner:}\quad &h_{t+1}(g)\in\argmin_{h\in\CC_t(g)}\widehat L_{\mathrm{train}}(h),\\
\text{outer:}\quad &\min_g\ \E[\widehat L_{\mathrm{val}}(h_{t+1}(g))].
\end{aligned}
\label{eq:bilevel-main}
\end{equation}
The expectation is taken over candidate generation and evaluation, with the task sets fixed. The inner loop optimizes the executor harness $h$ for a given $g$; the outer loop updates the optimizer harness $g$ to improve how it generates and checks candidates \citep{zakerinia2024flexible,sucker2025learning}. The final harness is selected on $\Sval$, and $\Stest$ is used only to evaluate it.

\textbf{Verification: choosing the right repair.}
A tool call may fail because of a timeout or incorrect arguments, requiring different repairs. We model this as a binary hypothesis test \citep{nielsen2014generalized} with two possible causes $W\in\{1,2\}$, prior probabilities $\lambda$ and $1-\lambda$, and repairs $h_1,h_2$; $h_W$ is the repair for the actual cause. The optimizer does not know $W$: it examines the available evidence, runs checks, and selects a repair $\widehat h$. The overlap of a piece of evidence is the Bhattacharyya coefficient \citep{bhattacharyya1943measure} between its distributions under the two causes: $1$ if it is equally likely under both, $0$ if it reveals the cause. Let $\beta_0$ be the overlap of all evidence available before the checks, such as the recorded traces, earlier feedback, and notes. Let $\rho_j$ bound the overlap of check $j$, which may depend on earlier results (Appendix~\ref{app:diagnosis}).

\begin{theorem}[Repair choice]
\label{thm:diagnosis}
Let $p_n^*$ be the smallest probability of choosing the wrong repair that any decision rule achieves from the evidence after $n$ checks, and let $\eps\ge0$ be the optimizer's excess error over this best rule. Then
\begin{equation}
\Prb(\widehat h\ne h_W)=p_n^*+\eps\le\tfrac12\,\beta_0\prod_{j=1}^{n}\rho_j+\eps.
\label{eq:diagnosis-main}
\end{equation}
If the checks add no information about the cause beyond the existing evidence, as when they only re-analyze it, then $p_n^*\ge\lambda(1-\lambda)\,\beta_0^2$ for every $n$.
\end{theorem}
\begin{remark}
The two bounds motivate verifying an edit before it is submitted. Re-analyzing the existing evidence cannot lower the smallest achievable error below $\lambda(1-\lambda)\beta_0^2$, even for a self-evolving optimizer; new training results between rounds add evidence for every method. Diagnoses from reading are often wrong in practice; in our replay study, an LLM reading the traces names the injected faulty step or a step next to it only 57.3\% of the time (Appendix~\ref{app:replay}). Verification adds execution evidence before submission, and a check whose feedback depends on the cause can lower the overlap and tighten the upper bound (Appendix~\ref{app:diagnosis}). Self-evolution can act on the remaining terms by choosing more informative checks (lower $\rho_j$) and using their evidence better (lower $\eps$).
\end{remark}

\Needspace{12\baselineskip}
\textbf{Self-evolution: finding and selecting a good harness.}
Theorem~\ref{thm:diagnosis} concerns how to repair a given failure. The optimizer must also decide which failure to repair, and the training and validation losses must then select a good candidate. Fix a target loss $\ell_0\in[0,1]$ and call a harness \emph{good} if its loss on new tasks is at most $\ell_0$. A run makes $K=TM$ proposals, $M$ candidates in each of $T$ rounds. Let $\CC$ contain $h_0$ and all $K$ candidates, and let $\eps_{\mathrm{sel}}=\E[L_{\theta,D}(\widehat h)-\min_{h\in\CC}L_{\theta,D}(h)]$ be the selection error, the expected loss gap between the selected harness $\widehat h$ and the best harness in $\CC$.

\begin{theorem}[Finding and selecting a good harness]
\label{thm:search}
Suppose that, as long as no good harness has been found, proposal $i$ targets a failure whose correct repair yields a good harness with probability at least $r_i$. Given such a target, suppose it submits a wrong repair with probability at most $e_i$. Then
\begin{equation}
\E L_{\theta,D}(\widehat h)\le \ell_0+
\underbrace{(1-\ell_0)\prod_{i=1}^{K}\bigl[1-r_i(1-e_i)\bigr]}_{\text{no good harness found}}
+\underbrace{\eps_{\mathrm{sel}}}_{\text{selection error}}.
\label{eq:search-main}
\end{equation}
\end{theorem}
\begin{remark}
Until a good harness is found, each proposal finds one with probability at least $r_i(1-e_i)$, since it must both target the right failure and repair it correctly. Verification can lower $e_i$ when its checks separate the causes and the optimizer uses the evidence well (Theorem~\ref{thm:diagnosis}; Appendix~\ref{app:bridge}). Self-evolution revises $g$ between rounds and can raise $r_i$ and lower $e_i$ in later rounds. The two factors multiply, so in the bound, better target selection helps little while repairs often fail (Appendix~\ref{app:bridge}). This is consistent with Observation~1, where self-evolution without verification did not improve validation accuracy. Finally, finding a good harness is not enough, because training and validation losses must also select it; Appendix~\ref{app:screening} bounds this selection error.
\end{remark}

\section{Experiments}
\label{sec:experiments}

\subsection{Setup}
\label{sec:setup}

We use the SWE-rebench splits of Section~\ref{sec:splits}: all training, validation, and in-distribution test tasks are in Python, while 87 of the 107 OOD test tasks are in Go, Java, Rust, or TypeScript. We evaluate every harness on both test sets. As in Observation~1, the optimizer is Qwen3.8-Flash-Next, the executor is Qwen3.8-27B, and each run has six rounds. Every method generates three candidate executor harnesses per round with the same proposal-turn limit; the verification tools account for 2 to 7\% of all executor runs during evolution, and total compute differs across methods (Appendix~\ref{app:cost}). We also run experiments on Terminal-Bench~\citep{terminalbench2026}; Appendix~\ref{app:details} gives its splits and setup, and Appendix~\ref{app:closed} its results.

We reimplement Meta-Harness~\citep{metaharness2026}, AHE~\citep{ahe2026}, Self-Harness~\citep{selfharness2026}, and HarnessX~\citep{harnessx2026} under this protocol, keeping each method's evidence interface (Appendix~\ref{app:baselines}), and add \method{} to each as its host. We report avg@3 accuracy: the percentage of test tasks resolved, averaged over three evaluations of a harness, with the standard error (SEM). Blank, the initial harness $h_0$, is evaluated the same way. Every run starts from $h_0$ and a new optimizer workspace.

\subsection{Main results}
\label{sec:main-results}

\begin{table}[t]
\caption{\textbf{Test accuracy of \method{} on SWE-rebench} (avg@3 accuracy, \%, $\pm$SEM over three test evaluations; one evolution run per row). In-distribution: the sealed test set (108 tasks); out-of-distribution (OOD): 107 newer tasks in five languages, tested with the same harnesses. $r^{\star}$: the round selected on validation; \emph{Val-selected}: the harness of that round; \emph{Last}: the final-round harness. At $r^{\star}=0$ the entry re-evaluates $h_0$. \textbf{Bold}: higher mean than the baseline of its block.}
\label{tab:equip-open}
\vspace{7pt}
\centering
\small
\setlength{\tabcolsep}{4pt}
\renewcommand{\arraystretch}{1.05}
\begin{tabular}{l c cc cc}
\toprule
& & \multicolumn{2}{c}{\textbf{In-distribution}} & \multicolumn{2}{c}{\textbf{Out-of-distribution}} \\
\cmidrule(lr){3-4} \cmidrule(lr){5-6}
Method & $r^{\star}$ & Val-selected & Last & Val-selected & Last \\
\midrule
Blank (initial harness $h_0$) & -- & $33.95_{\pm0.82}$ & $33.95_{\pm0.82}$ & $27.41_{\pm0.82}$ & $27.41_{\pm0.82}$ \\
\midrule
Meta-Harness~\citep{metaharness2026} & 4 & $39.20_{\pm1.11}$ & $35.80_{\pm0.82}$ & $28.66_{\pm0.82}$ & $30.84_{\pm1.95}$ \\
\quad $+$ \method{} (w/o self-evolve) & 6 & $34.57_{\pm0.31}$ & $34.57_{\pm0.31}$ & $28.35_{\pm1.36}$ & $28.35_{\pm1.36}$ \\
\quad $+$ \method{} (w/ self-evolve) & 2 & $\mathbf{42.28}_{\pm1.11}$ & $\mathbf{41.36}_{\pm2.16}$ & $\mathbf{37.69}_{\pm1.56}$ & $\mathbf{38.01}_{\pm1.56}$ \\
\addlinespace[1pt]
\arrayrulecolor{black!50}\hdashline[1.8pt/2.2pt]
\arrayrulecolor{black}\addlinespace[2pt]
AHE~\citep{ahe2026} & 6 & $33.95_{\pm1.23}$ & $33.95_{\pm1.23}$ & $26.17_{\pm0.54}$ & $26.17_{\pm0.54}$ \\
\quad $+$ \method{} (w/o self-evolve) & 2 & $\mathbf{37.96}_{\pm2.14}$ & $32.10_{\pm0.31}$ & $\mathbf{35.51}_{\pm1.08}$ & $\mathbf{34.58}_{\pm1.95}$ \\
\quad $+$ \method{} (w/ self-evolve) & 5 & $\mathbf{35.19}_{\pm2.14}$ & $\mathbf{35.80}_{\pm1.11}$ & $\mathbf{30.22}_{\pm0.62}$ & $\mathbf{30.22}_{\pm1.89}$ \\
\addlinespace[1pt]
\arrayrulecolor{black!50}\hdashline[1.8pt/2.2pt]
\arrayrulecolor{black}\addlinespace[2pt]
Self-Harness~\citep{selfharness2026} & 6 & $29.63_{\pm0.53}$ & $29.63_{\pm0.53}$ & $29.28_{\pm2.72}$ & $29.28_{\pm2.72}$ \\
\quad $+$ \method{} (w/o self-evolve) & 5 & $\mathbf{40.12}_{\pm1.35}$ & $\mathbf{40.12}_{\pm1.63}$ & $\mathbf{31.78}_{\pm2.47}$ & $\mathbf{35.20}_{\pm0.31}$ \\
\quad $+$ \method{} (w/ self-evolve) & 1 & $\mathbf{34.88}_{\pm1.35}$ & $\mathbf{33.33}_{\pm1.93}$ & $\mathbf{30.22}_{\pm0.82}$ & $\mathbf{31.78}_{\pm1.95}$ \\
\addlinespace[1pt]
\arrayrulecolor{black!50}\hdashline[1.8pt/2.2pt]
\arrayrulecolor{black}\addlinespace[2pt]
HarnessX~\citep{harnessx2026} & 0 & $32.72_{\pm1.11}$ & $34.26_{\pm2.45}$ & $27.41_{\pm1.25}$ & $32.09_{\pm1.65}$ \\
\quad $+$ \method{} (w/o self-evolve) & 1 & $\mathbf{38.89}_{\pm2.45}$ & $32.10_{\pm1.35}$ & $27.10_{\pm1.43}$ & $29.28_{\pm0.31}$ \\
\quad $+$ \method{} (w/ self-evolve) & 3 & $\mathbf{35.80}_{\pm0.82}$ & $\mathbf{36.11}_{\pm1.60}$ & $\mathbf{37.69}_{\pm2.43}$ & $\mathbf{32.71}_{\pm0.93}$ \\
\bottomrule
\end{tabular}
\end{table}

\textbf{\method{} improves every baseline on both test sets.} With self-evolution, \method{} scores above its unequipped baseline in all 16 reported comparisons of Table~\ref{tab:equip-open}, on both test sets and for both the selected and the final harness, by $0.6$ to $10.3$ points. Its best harness, on the Meta-Harness host, is $3.1$ points above the strongest baseline in distribution. The four hosts analyze failed trajectories in different ways (Appendix~\ref{app:baselines}), and \method{} improves each of them, so it is not tied to one particular form of failure analysis.

\textbf{The gains of \method{} generalize out of distribution.} On the OOD test set, the validation-selected baselines stay within two points of blank, including Meta-Harness, whose gain over blank falls from 5.2 points in distribution to 1.2. With self-evolution, \method{} is 2.8 to 10.3 points above blank on every host. What \method{} learns on Python tasks thus carries over to Go, Java, Rust, and TypeScript: across the five languages, it raises its host in 16 of the 20 pairs of host and language (Appendix~\ref{app:languages}).

\textbf{Self-evolution keeps \method{} above the baseline in every comparison.} The \method{} tools already help without self-evolution: in distribution, they select a better harness than the baseline on three of the four hosts, by $4.0$ to $10.5$ points. Without self-evolution, however, they fall below the baseline in 8 of the 16 comparisons of Table~\ref{tab:equip-open}, while \method{} with self-evolution stays above it in all 16. Self-evolution does not raise every score: compared with \method{} without self-evolution, it raises the selected accuracy on Meta-Harness on both test sets and on HarnessX out of distribution, and lowers it in the other five cases. Its selected and final harnesses stay within $1.6$ points in distribution, but differ by $5.0$ points on HarnessX out of distribution. Test accuracy is not available when one chooses a host or a round, so staying above the baseline in every comparison matters in practice.

\subsection{Ablations}
\label{sec:ablation}

\begin{table}[t]
\caption{\textbf{Ablating \method{} on the AHE host} (SWE-rebench in-distribution test set; avg@3 accuracy, \%, $\pm$SEM over three test evaluations; one evolution run per row). Every configuration runs without self-evolution. Each indented row removes components from the full configuration; the last row is AHE alone. $r^{\star}$, \emph{Val-selected}, and \emph{Last} are as in Table~\ref{tab:equip-open}.}
\label{tab:ablation-open}
\vspace{5pt}
\centering
\small
\setlength{\tabcolsep}{4pt}
\renewcommand{\arraystretch}{1.05}
\begin{tabular}{l c cc}
\toprule
Configuration & $r^{\star}$ & Val-selected & Last \\
\midrule
AHE $+$ \method{} (w/o self-evolve), full configuration & 2 & $37.96_{\pm2.14}$ & $32.10_{\pm0.31}$ \\
\quad w/o verification during evolution (minimization kept) & 3 & $37.65_{\pm1.35}$ & $35.49_{\pm0.31}$ \\
\quad w/o attribution and minimization & 2 & $35.80_{\pm1.35}$ & $31.79_{\pm2.53}$ \\
\quad w/o training audit only & 2 & $36.73_{\pm1.23}$ & $32.41_{\pm3.21}$ \\
AHE without any \method{} component & 6 & $33.95_{\pm1.23}$ & $33.95_{\pm1.23}$ \\
\bottomrule
\end{tabular}
\par\vspace{9pt}
\includegraphics[width=\textwidth]{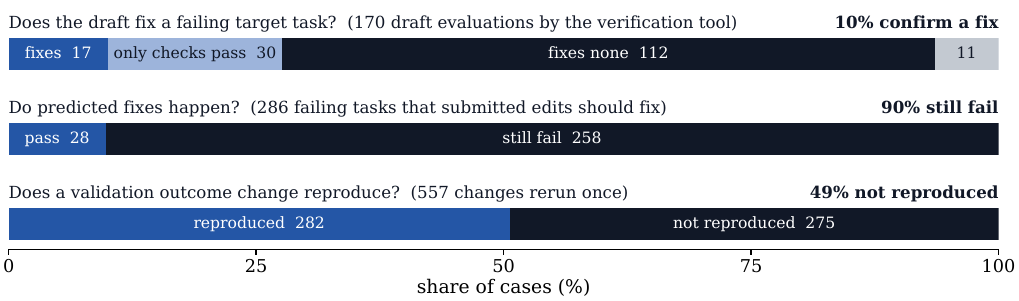}
\par\vspace{-3pt}
\normalsize
\captionof{figure}{\textbf{What verification does during evolution}, over the 18 runs of Tables~\ref{tab:observations}--\ref{tab:ablation-open}. Top: draft evaluations in the ten runs with the \texttt{verification} tool (light blue: only regression checks passed; gray: inconclusive). Middle: failing training tasks that submitted edits were predicted to fix, checked in the next training run. Bottom: validation outcome changes, rerun once.}
\label{fig:mechanism}
\vspace{-8pt}
\end{table}

\textbf{The diagnosis reports make the checks informative.} In Table~\ref{tab:ablation-open}, removing attribution and minimization, which also removes the training audit, costs about half of the gain that \method{} adds to AHE ($2.2$ of $4.0$ points); removing the training audit alone costs $1.2$. Without the reports, the optimizer calls the verification tools more than three times as often, yet selects a weaker harness. Execution checks likely help most when they test well-chosen hypotheses, which the reports supply.

\textbf{Without self-evolution, the optimizer uses the verification tools less.} Removing them barely changes the selected-round accuracy: on this host, the fixed optimizer's first nine calls all refuted its drafts, and it stopped calling after the second round. Across the four hosts of Table~\ref{tab:equip-open}, self-evolving optimizers call the tools more than twice as often, and more than three times as often in the last three rounds. On Meta-Harness, \method{} without self-evolution scores below both the baseline and simple verification alone (Tables~\ref{tab:observations} and~\ref{tab:equip-open}). Its optimizer kept calling the \texttt{verification} tool, but only on tasks that the current harness fails, so none of these calls could show whether an edit breaks a task that already passes. In 16 of the 23 drafts that the self-evolving optimizer ran on this host, the targets included such a task. In Appendix~\ref{app:ablations}, random or fixed experiment schedules and a restricted workspace also cost accuracy.

\subsection{What verification does during evolution}
\label{sec:mechanism}

\textbf{Few draft edits fix their targets, and execution shows it before submission.} Figure~\ref{fig:mechanism} aggregates the logs of the 18 runs behind Tables~\ref{tab:observations}--\ref{tab:ablation-open}. With each submitted edit, the optimizer names the failing training tasks it expects the edit to fix; only one in ten of these tasks passed in the next training run. The \texttt{verification} tool reveals such failures before submission: only 17 of its 170 draft evaluations, counting repeated evaluations of the same draft, found that the draft fixed a failing task it targeted (Appendix~\ref{app:mechanism}). In another 30, the draft passed only on tasks that the current harness already solves, which the optimizer had added to check that the draft does not break them. Such regression checks matter because an edit can fix its targets and still break other tasks: across the 18 runs, submitted edits fixed 148 validation tasks and broke 134, counting only changes that reproduced on a rerun. Half of all validation outcome changes did not reproduce. Because single runs are this noisy, the tool counts a task as fixed only after two passes.

\textbf{Self-evolution changes how the optimizer verifies.} Self-evolving optimizers learned to add regression checks before submission (Appendix~\ref{app:examples}): 34 of their 81 draft evaluations included a training task that the current harness already solves, against 2 of 89 for fixed optimizers. Self-evolving optimizers also respond when a check fails: when the tool showed that a draft fixed none of its targets, they revised the draft before submitting in 30 of 38 cases ($79\%$), against 42 of 74 ($57\%$) without self-evolution. Evaluated drafts fixed a failing target at similar rates in both groups (9 of 81 against 8 of 89); the clearer difference lies in what the optimizer tests and how it responds when a test fails.

\section{Conclusion}
\label{sec:conclusion}

\method{} combines execution-based verification with self-evolution of
the harness optimizer, while keeping both models' weights fixed.
The optimizer can test draft edits, replay failures, and perturb
suspected steps before submission, then use the resulting evidence
to revise both the executor harness and its own prompts, skills,
tools, hooks, and notes. Our analysis identifies conditions under
which informative checks can reduce repair-choice error and relates
downstream performance to the quality of target selection, repair,
and harness selection. In our experiments, adding \method{} to four reimplemented
harness optimizers improves all four on held-out Python tasks and
on newer out-of-distribution tasks spanning five programming languages.
The evolution logs reveal how the optimization process changes:
self-evolving optimizers more often include regression checks and
revise unsuccessful drafts before submission. Although the incremental
benefit of self-evolution varies across hosts, these findings support
a broader principle: harness evolution should improve not only
the agent's configuration, but also the procedures that propose,
test, and revise it.

\subsection*{AI use statement}

In this work, we used generative AI tools to draft and edit the manuscript, search and check related work and citations, implement experiment infrastructure, help analyze experiment logs, and help write and check the theoretical results and proofs. Benchmark tasks come from existing datasets; the trajectories, diagnoses, edits, and optimizer artifacts that we analyze are outputs of the models under study. The research ideas, the ideas and structure of the proofs, the experimental design, and the interpretation of the analyses are the authors' own. All AI-assisted text, code, analyses, and proofs were reviewed, verified, and tested by the authors. We take responsibility for the final content of this work, including text, claims, and artifacts produced with the aid of generative AI.

\subsection*{Ethics statement}

This work studies methods for improving LLM agent harnesses on software engineering benchmarks in sandboxed environments. It does not involve human subjects or sensitive data. Self-modifying optimizers raise oversight questions. In our system, the internals of the verification tools are fixed, all optimizer actions are logged and auditable, self-written code is loaded only after a safety check and a test run, the executor acts only inside task containers without network access, and the executor model is frozen, which bounds the scope of self-modification.

\subsection*{Reproducibility statement}

Section~\ref{sec:splits} and Appendix~\ref{app:details} specify the task pools, splits, and budgets; Section~\ref{sec:setup} specifies the models and the evaluation protocol; Section~\ref{sec:overview} specifies the verification tools and optimizer self-evolution; Appendix~\ref{app:algorithm} gives pseudocode, including the selection rules; Appendix~\ref{app:baselines} describes the baseline reimplementations; and Appendix~\ref{app:theory} states the theoretical assumptions and proofs. Our repository at \url{https://github.com/wzekai/VERSE} contains the code (the evolution engine, the verification tools, the baseline reimplementations, and the evaluation scripts), the configuration of every reported evolution run, and the task IDs of every split; for the main results table, it also contains the selected executor and optimizer harnesses and their per-task test results.

\bibliography{verso}
\bibliographystyle{iclr2027_conference}

\clearpage
\appendix
\raggedbottom

\etocdepthtag.toc{mtappendix}
\etocsettagdepth{mtchapter}{none}
\etocsettagdepth{mtappendix}{subsection}
\makeatletter
\renewcommand*\l@section[2]{%
  \ifnum \c@tocdepth >\z@
    \addpenalty\@secpenalty
    \addvspace{0.45em \@plus\p@}%
    \setlength\@tempdima{1.5em}%
    \begingroup
      \parindent \z@ \rightskip \@pnumwidth
      \parfillskip -\@pnumwidth
      \leavevmode \bfseries
      \advance\leftskip\@tempdima
      \hskip -\leftskip
      #1\nobreak\hfil
      \nobreak\hb@xt@\@pnumwidth{\hss #2\kern-\p@\kern\p@}\par
    \endgroup
  \fi}
\makeatother
\etocsettocstyle{\section*{Appendix contents}}{}
\tableofcontents
\clearpage

\section{Related work}
\label{app:related}

\textbf{Harness evolution.}
Recent methods improve an agent's harness with an LLM proposer that reads execution traces and scores; our four baselines differ in how the proposer reads this evidence~\citep{metaharness2026,ahe2026,selfharness2026,harnessx2026} (Appendix~\ref{app:baselines}). Related work adapts harnesses at test time~\citep{tthe2026}, refines prompt-level harnesses from pairwise comparisons~\citep{rhi2026}, optimizes prompts and programs with LLM feedback~\citep{shinn2023reflexion,yang2024opro,khattab2024dspy}, updates model weights~\citep{sia2026}, or distills skills from trajectories~\citep{trace2skill2026,evoskill2026}. \citet{rethinking2026} note that many results search and report on the same tasks; ModularRSI evolves on tasks disjoint from its benchmarks~\citep{wu2026modularrsi}, and RRSI regularizes proposal and selection and reports out-of-distribution gains~\citep{xia2026rrsi}.

\textbf{Verifying harness edits.}
Execution is also used to screen candidate harnesses with gates~\citep{harnessbank2026}, repair attributed failures~\citep{harnessfix2026}, and revert and replay agent runs~\citep{shepherd2026}. SPEAR gives a prompt optimizer an evaluate tool that it calls itself~\citep{spear2026}, and VeRO finds that agent optimizers mostly edit prompts~\citep{vero2026}. Among contemporaneous systems, AutoSaddler keeps a patch only if it improves a mini-batch, then evaluates it on a development set~\citep{autosaddler2026}, HarnessEvolve gates updates and locates errors against reference trajectories~\citep{jiang2026harnessevolve}, and EnvHarness accepts environment components by success rates over fresh rollouts~\citep{envharness2026}.

\textbf{Evolving the optimizer.}
In most harness optimizers, the optimizer's own procedure stays fixed during a run; when the optimizer and the executor share a base model, the optimizer runs under its own prompts or scaffold~\citep{ahe2026,tthe2026}. In Self-Harness and HASE, the model proposes edits under the harness it is editing, so one harness serves both roles~\citep{selfharness2026,hase2026}. In self-referential agents, one program both solves tasks and improves itself: STOP runs a program improver on its own code~\citep{stop2024}, G{\"o}del Agent modifies its own logic~\citep{yin2025godel}, the Darwin G{\"o}del Machine lets coding agents modify their own code~\citep{dgm2026}, and Hyperagents place a task agent and a meta agent in one editable program, so the meta agent can rewrite itself~\citep{zhang2026hyperagents}.

Other systems keep the optimizer separate and revise its procedure. Some add a meta-level above it: a meta-evolution loop over the evolution blueprint~\citep{seong2026lastharness}, a meta-agent that edits the procedure controlling future evolution~\citep{zhang2026aevo} or a base agent's context-engineering skills~\citep{mce2026}, a meta-evolver that improves a harness evolver's strategy~\citep{zhou2026hsi}, and a meta-level policy that schedules improvement operators~\citep{tan2026metarsi}. In others, the optimizer revises its own procedure. Promptbreeder evolves the mutation prompts that improve its task prompts~\citep{fernando2024promptbreeder}, SePO optimizes a prompt agent's own system prompt~\citep{tao2026sepo}, MetaSkill-Evolve applies its improvement pipeline to its own meta-skill~\citep{wang2026metaskillevolve}, and EvoTrainer's trainer revises its own harness across training runs~\citep{evotrainer2026}.

\textbf{Position of \method{}.}
Like the last group, the \method{} optimizer revises its own procedure, but its harness $g$ is a full agent harness kept separate from the executor harness $h$: it rewrites its own prompts, skills, notes, hooks, and executable tools. While editing $h$, it calls verification tools that test the causal claim behind an edit before submission; AutoSaddler, HarnessEvolve, and EnvHarness also verify edits, with fixed optimization procedures. The executor model stays frozen and is evaluated on disjoint task splits. In our controlled study, self-evolution lowered accuracy without verification and gave the best result with it (Section~\ref{sec:two-observations}).

\section{Pseudocode for the bilevel learning protocol}
\label{app:algorithm}
A harness is a complete executable configuration, including prompts, tools, control flow, dependencies, and persistent state. The executor weights $\theta$ and optimizer weights $\omega$ remain fixed; $h$ controls task execution and $g$ controls candidate generation and checking. All tasks, trajectories, harnesses, and evidence records have finite encodings. Their spaces are countable, with all subsets measurable; $\HH$ denotes the space of executor harnesses. Learning $g$ treats the optimizer as a learned algorithm \citep{zakerinia2024flexible}, using execution feedback for program revision \citep{agrawal2026gepa}.

The inner loop compares the nonempty candidate set $\CC_t(g)$ on the same training tasks; if a prescribed training subset is used, the comparison uses its mean loss. The outer objective conditions on the current state and task sets and averages over candidate generation and evaluation. Candidates may run in parallel, and their results need not be independent.

\begin{algorithm}[t]
\caption{Verified self-evolving harness learning}
\label{alg:bilevel}
\begin{algorithmic}[1]
\Require Fixed executor weights $\theta$ and optimizer weights $\omega$; training $\Str$,
\Statex \hspace{\algorithmicindent}validation $\Sval$, and test $\Stest$; initial $h_0,g_0$;
\Statex \hspace{\algorithmicindent}round count $T\ge1$, candidates per round $M\ge1$, budgets $B_t$
\Ensure Frozen executor harness $\widehat h$ and its test evaluation
\State Evaluate $h_0$ on $\Sval$ and archive the harness and result
\For{$t=0,\ldots,T-1$} \Comment{Outer loop: improve the optimizer}
    \State Fix $g_t$; $\CC_t\gets\varnothing$; training record $E_t\gets\varnothing$
    \For{$c=1,\ldots,M$} \Comment{Inner loop: search for executor harnesses}
        \State $\widetilde h\gets h_t$
        \While{the candidate's turn limit is not reached}
            \State Read training traces, earlier validation feedback $F_{<t}$, and prior evidence under $g_t$
            \State Propose an edit to $\widetilde h$ or a diagnostic check on $\Str$
            \If{a check is requested and the round's shared budget remains}
                \State Execute the check; append raw outcomes and costs to $E_t$
            \EndIf
            \State Revise $\widetilde h$ using the available evidence
        \EndWhile
        \State $\CC_t\gets\CC_t\cup\{\widetilde h\}$
    \EndFor
    \State Evaluate $\CC_t$ on the training comparison tasks
    \State Choose $h_{t+1}\in\argmin_{h\in\CC_t}\widehat L_{\mathrm{train}}(h)$, or keep $h_t$ if $\CC_t=\varnothing$
    \State Smoke-check $h_{t+1}$ and let the optimizer repair it if needed; record results in $E_t$
    \State Evaluate $h_{t+1}$ on $\Sval$ to obtain feedback $F_t$; archive both
    \State Use the optimizer to revise $g_t$ from $E_t,F_t$, obtaining $g_{t+1}$
\EndFor
\State Re-evaluate the two archived harnesses with the lowest validation losses; keep the better as $\widehat h$
\State Freeze $\widehat h$ and evaluate it on $\Stest$
\State \Return $\widehat h$ and its test evaluation
\end{algorithmic}
\end{algorithm}

\textbf{Evidence and feedback.} $E_t$ contains the training checks and comparison results of round $t$. $F_t$ is the validation feedback returned to the optimizer, the same for every method: the validation loss of each round, the identifiers of up to 30 validation tasks whose outcome changed and whether each change held on a rerun, and the rate and two most frequent messages of harness-code errors on validation runs. It never contains task statements, tests, or trajectories. Test tasks enter neither loop.

\textbf{Checks.} All candidates and the optimizer self-update draw checks from one execution budget per round, and every check runs on training tasks. A check that cannot run, because the round's budget or time is spent or the draft cannot be applied, returns an inconclusive result; a draft that cannot be applied costs no budget. No check result blocks a submission: the optimizer decides whether to revise, check again, or submit.

\textbf{Choosing the harness of a round.} The candidates are compared by the net number of training tasks they fix on a shared subset of up to 12 training tasks. A tie goes to the candidate whose submitted edit has more tasks that the \texttt{verification} tool confirmed as passing, and then to a fixed order. The chosen candidate must pass a smoke check that includes one run on a training task. If it fails, the optimizer repairs it in at most three attempts, and the repaired version is validated and archived as $h_{t+1}$ without being screened again. If no attempt passes, or if no candidate was submitted, the round keeps $h_t$ and skips validation; the self-update still runs.

\textbf{Validation and final selection.} A validation task whose outcome differs from that of the parent harness is rerun once, and a change that does not reproduce is reverted to the parent's outcome. At the end, the two archived harnesses with the lowest validation losses are each evaluated once more, and the one with the lower mean loss is selected; ties favor the later round. Appendix~\ref{app:screening} analyzes this rule and the reliability of validation losses under feedback.

\textbf{Self-updates.} A self-update of the optimizer harness takes effect only if its code passes a safety check and a test run; otherwise the previous version stays in place.

\section{Additional theoretical analysis}
\label{app:theory}
This appendix gives the formal versions of Theorems~\ref{thm:diagnosis} and~\ref{thm:search} and the analysis of selection. Tasks, trajectories, harnesses, and evidence have finite encodings, so all spaces are countable and all subsets measurable. Logarithms are natural.

\textbf{Tools of the optimizer.} All methods in our experiments share the basic tools $\mathcal T_{\mathrm{basic}}$ of \eqref{eq:toolkit-theory}. They read and search trajectories, run shell commands for analysis, write notes that carry over to later rounds, and submit proposed edits to the executor harness. Attribution and the training audit are not in $\mathcal T_0$; they run automatically each round and add their reports to the optimizer's evidence. Self-evolving optimizers also keep simple verification (Section~\ref{sec:two-observations}), which the \texttt{verification} tool generalizes to up to three tasks; their self-written tools can call both. Later optimizer harnesses keep all tools in $\mathcal T_0$, can add new tools built from them, and can revise the other parts of $g$.

\subsection{Choosing a repair from verification evidence}
\label{app:diagnosis}
\subsubsection{Checks as an adaptive hypothesis test}
Fix two distinct repairs $h_1,h_2$ for two possible causes $W\in\{1,2\}$ with prior probabilities $\lambda,1-\lambda$, $\lambda\in[0,1]$. The optimizer starts from the evidence $\mathcal I_0$ gathered before any check, such as trajectories, attribution reports, earlier feedback, and notes, and may make up to $n$ checks. Check $j$ consists of a query $Q_j$ and feedback $Y_j$:
\begin{equation}
Q_j\sim\Pi_j(\cdot\mid\mathcal I_{j-1}),\qquad
Y_j\sim \kappa_W(\cdot\mid\mathcal I_{j-1},Q_j),\qquad
\mathcal I_j=(\mathcal I_{j-1},Q_j,Y_j).
\label{eq:channel}
\end{equation}
The query policy $\Pi_j$ belongs to the optimizer. It may depend on the whole history, including revisions of the optimizer harness, but not on $W$, which the optimizer cannot observe. The feedback law $\kappa_w$ belongs to the sandbox and may differ between causes. A run that stops early makes null checks with constant feedback until $n$; they carry no information.

For probability masses $P,P'$ on a countable set, the Bhattacharyya coefficient is $\Aff(P,P')=\sum_y\sqrt{P(y)P'(y)}$~\citep{bhattacharyya1943measure}. It lies in $[0,1]$, equals $1$ for identical distributions, and equals $0$ for disjoint supports. The coefficient of check $j$ is
\begin{equation}
\rho_j=\sup_{\mathcal I}\sum_q\Pi_j(q\mid\mathcal I)\,
\Aff\!\left(\kappa_1(\cdot\mid\mathcal I,q),\kappa_2(\cdot\mid\mathcal I,q)\right),
\label{eq:rho}
\end{equation}
where the supremum runs over histories at step $j-1$ with positive probability under both causes, and $\rho_j=0$ if there are none. It averages the ambiguity of the feedback over the checks the optimizer chooses at each history. The sandbox determines $\kappa_1,\kappa_2$ and the optimizer determines $\Pi_j$; either can make $\rho_j$ small.

\subsubsection{Proof of Theorem~\ref{thm:diagnosis}}
\begin{lemma}[Bayes error and the Bhattacharyya coefficient]
\label{lem:binary}
Let $P_1,P_2$ be the laws of the evidence under the two causes and $\beta=\Aff(P_1,P_2)$. The smallest error probability of any decision rule, $p^*=\sum_{\mathcal I}\min\{\lambda P_1(\mathcal I),(1-\lambda)P_2(\mathcal I)\}$, satisfies
\[
\lambda(1-\lambda)\beta^2\le p^*\le\sqrt{\lambda(1-\lambda)}\,\beta\le\tfrac12\beta.
\]
\end{lemma}
The upper bound is the Bhattacharyya bound on the Bayes error and the lower bound is due to \citet{kailath1967divergence}; \citet{nielsen2014generalized} presents both.
\begin{proof}
The rule that picks the cause with the larger posterior mass at each $\mathcal I$ leaves the smaller mass as its error, and no rule, randomized or not, does better pointwise. Write $a=\lambda P_1(\mathcal I)$ and $b=(1-\lambda)P_2(\mathcal I)$, so that $\sum\min(a,b)=p^*$, $\sum\max(a,b)=1-p^*$, and $\sum\sqrt{ab}=\sqrt{\lambda(1-\lambda)}\,\beta$. The upper bounds follow from $\min(a,b)\le\sqrt{ab}$ and $\lambda(1-\lambda)\le\tfrac14$. For the lower bound, $\sqrt{ab}=\sqrt{\min(a,b)\max(a,b)}$ and the Cauchy--Schwarz inequality give $\lambda(1-\lambda)\beta^2\le p^*(1-p^*)\le p^*$.
\end{proof}

\begin{lemma}[Each check multiplies the coefficient]
\label{lem:adaptive}
Let $P_w^{(j)}$ be the law of $\mathcal I_j$ under cause $w$ and $\beta_j=\Aff(P_1^{(j)},P_2^{(j)})$. Then $\beta_j\le\rho_j\beta_{j-1}$, and $\beta_j=\beta_{j-1}$ when $\kappa_1=\kappa_2$ at check $j$.
\end{lemma}
\begin{proof}
Under \eqref{eq:channel}, $P_w^{(j)}(\mathcal I,q,y)=P_w^{(j-1)}(\mathcal I)\,\Pi_j(q\mid\mathcal I)\,\kappa_w(y\mid\mathcal I,q)$, and the factor $\Pi_j$ is the same under both causes. Hence
\begin{align}
\beta_j
&=\sum_{\mathcal I}\sqrt{P_1^{(j-1)}(\mathcal I)P_2^{(j-1)}(\mathcal I)}
\sum_q\Pi_j(q\mid\mathcal I)\sum_y\sqrt{\kappa_1(y\mid\mathcal I,q)\kappa_2(y\mid\mathcal I,q)}
\le\rho_j \beta_{j-1},
\label{eq:recursion}
\end{align}
because the inner double sum is at most $\rho_j$ at every history with positive mass under both causes, and a history with zero mass under either cause contributes nothing. If $\kappa_1=\kappa_2$, the innermost sum equals $1$ and the expression equals $\beta_{j-1}$.
\end{proof}

\begin{formaltheorem}{\ref{thm:diagnosis}}
Let $p_n^*=\inf_\psi\Prb\{\psi(\mathcal I_n)\ne h_W\}$ over decision rules $\psi$ on complete histories. Then
\begin{equation}
p_n^*\le\sqrt{\lambda(1-\lambda)}\,\beta_0\prod_{j=1}^n\rho_j\le\tfrac12\,\beta_0\prod_{j=1}^n\rho_j,
\label{eq:diagnosis-formal}
\end{equation}
and if $\kappa_1=\kappa_2$ at every check, then $p_n^*\ge\lambda(1-\lambda)\beta_0^2$. The optimizer's choice $\widehat h$, which may use the history and private randomness, errs with probability $p_n^*+\eps$ for some $\eps\ge0$.
\end{formaltheorem}
\begin{proof}
Lemma~\ref{lem:adaptive} gives $\beta_n\le\beta_0\prod_j\rho_j$, with $\beta_n=\beta_0$ when no check depends on the cause. Lemma~\ref{lem:binary} applied to the laws of $\mathcal I_n$ gives both bounds, and $\eps\ge0$ because $p_n^*$ is the minimum over all rules.
\end{proof}

The lower bound holds for every query policy, hence for every optimizer harness. A check that only re-analyzes the recorded evidence, such as re-reading traces or running static analysis over them, has $\kappa_1=\kappa_2$ and leaves the coefficient unchanged. A check can lower it only if its feedback distinguishes the causes beyond the existing evidence; executing a draft repair provides such feedback. For example, \texttt{perturbation} can replace a command whose arguments may be wrong and replay the remaining steps.

The confirmation rerun of the \texttt{verification} tool lowers the coefficient further. Consider a check that runs the draft repair $h_1$ on a target training task, and let it pass with probability $q_1$ when cause $1$ is present and $q_2$ otherwise. The coefficient of this check is $\rho=\sqrt{q_1q_2}+\sqrt{(1-q_1)(1-q_2)}$. The \texttt{verification} tool reruns the task after a pass. With independent runs the outcomes are a fail, a pass followed by a fail, and two passes, and the coefficient becomes
\[
\sqrt{(1-q_1)(1-q_2)}+\sqrt{q_1q_2}\,\rho=\rho-\sqrt{q_1q_2}\,(1-\rho),
\]
which is below $\rho$ whenever $q_1q_2>0$ and $\rho<1$. A wrong repair that passes by chance with probability $q_2$ passes twice with probability $q_2^2$.

\subsection{Finding a good harness}
\label{app:search}
\subsubsection{Conditions and proof of Theorem~\ref{thm:search}}
Write $L(h)=L_{\theta,D}(h)$ for a fixed executor and task distribution; expectations run over the whole run. The run makes $K$ proposals in a fixed order; $K=TM$ in Algorithm~\ref{alg:bilevel}, where the candidates of one round may run in parallel and their outcomes may be dependent. Let $h^{(i)}$ be the harness submitted by proposal $i$, repaired if it fails the smoke check (Appendix~\ref{app:algorithm}), and let $\CC=\{h_0,h^{(1)},\ldots,h^{(K)}\}$. If proposal $i$ produces no candidate, $h^{(i)}$ is the current harness, so such failures only lower $r_i$. The selected harness $\widehat h$ lies in $\CC$.

Fix $\ell_0\in[0,1]$. Let $A_i$ be the event that $h_0$ and $h^{(1)},\ldots,h^{(i)}$ all have loss above $\ell_0$, with $A_0=\{L(h_0)>\ell_0\}$. Let $O_i$ be the event that proposal $i$ targets a failure whose correct repair has loss at most $\ell_0$, and $J_i$ the event that it submits that repair, so that $O_i\cap J_i\subseteq\{L(h^{(i)})\le\ell_0\}$. The conditions of Theorem~\ref{thm:search} are, for constants $r_i,e_i\in[0,1]$,
\begin{equation}
\Prb(O_i\mid A_{i-1})\ge r_i,\qquad
\Prb(J_i^c\mid O_i,A_{i-1})\le e_i,
\label{eq:opportunity}
\end{equation}
whenever the conditioning events have positive probability; if $\Prb(O_i\mid A_{i-1})=0$ we set $r_i=0$. These conditions allow proposals to depend on earlier outcomes and on revisions of the optimizer harness; independence between proposals is not required. The repair must lower the population loss, not only fix the training task it was found on.

\begin{formaltheorem}{\ref{thm:search}}
Under \eqref{eq:opportunity}, with $\eps_{\mathrm{sel}}=\E[L(\widehat h)-\min_{h\in\CC}L(h)]$,
\begin{equation}
\E L(\widehat h)\le
\ell_0+(1-\ell_0)\prod_{i=1}^K[1-r_i(1-e_i)]+\eps_{\mathrm{sel}}.
\label{eq:search-formal}
\end{equation}
If $r_i\ge r$ and $e_i\le e$ for all $i$, the middle term is at most $(1-\ell_0)[1-r(1-e)]^K$.
\end{formaltheorem}
\begin{proof}
When $\Prb(A_{i-1})>0$ and $\Prb(O_i\mid A_{i-1})>0$,
\[
\Prb\{L(h^{(i)})\le\ell_0\mid A_{i-1}\}
\ge\Prb(O_i\mid A_{i-1})\,\Prb(J_i\mid O_i,A_{i-1})\ge r_i(1-e_i),
\]
and the bound also holds when either probability is zero. Since $A_i=A_{i-1}\cap\{L(h^{(i)})>\ell_0\}$, this gives $\Prb(A_i)\le[1-r_i(1-e_i)]\Prb(A_{i-1})$, and iterating from $\Prb(A_0)\le1$,
\[
\Prb\Bigl\{\min_{h\in\CC}L(h)>\ell_0\Bigr\}=\Prb(A_K)\le\prod_{i=1}^K[1-r_i(1-e_i)].
\]
The minimum loss is at most $\ell_0$ off $A_K$ and at most $1$ on it, so $\E\min_{\CC}L\le\ell_0+(1-\ell_0)\Prb(A_K)$. Adding $\eps_{\mathrm{sel}}$ proves the claim.
\end{proof}

\subsubsection{Connecting verification to the repair error}
\label{app:bridge}
Theorem~\ref{thm:diagnosis} bounds $e_i$ when the repair decision of proposal $i$ is a test between two causes. Condition on $O_i$, $A_{i-1}$, and the proposal's context, and suppose that within this context the optimizer runs the adaptive test of Appendix~\ref{app:diagnosis} with at most $n_i$ checks and submits the repair it selects. Let $\rho_{i,j}$ bound the coefficient of check $j$ and $\eps_i$ the optimizer's excess error, uniformly over contexts. Applying Theorem~\ref{thm:diagnosis} in each context and averaging gives
\begin{equation}
\Prb(J_i^c\mid O_i,A_{i-1})\le
 e_i:=\min\!\left\{1,\frac12\prod_{j=1}^{n_i}\rho_{i,j}+\eps_i\right\},
\label{eq:bridge}
\end{equation}
which can be substituted into \eqref{eq:search-formal}.

If every proposal makes $n$ checks with coefficient at most $\rho<1$ and excess error at most $\eps$, then $e_i\le\min\{1,\tfrac12\rho^n+\eps\}$ for every $i$. The first term decays geometrically in the number of checks, so a few checks bring the bound close to $\eps$; further gains come from lower $\eps$ and higher $r_i$, which self-evolution can act on.

The two factors of $r_i(1-e_i)$ multiply. Increasing $r_i$ by $\Delta$ raises the lower bound on the success probability of a proposal by only $\Delta(1-e_i)$, so better target selection helps little while repairs often fail. Within each benchmark, all methods use the same $K$ (Section~\ref{sec:setup}), so in \eqref{eq:search-main} they differ only through $r_i$, $e_i$, and $\eps_{\mathrm{sel}}$.

\subsection{Selection error}
\label{app:screening}
\subsubsection{Two stages}
Theorem~\ref{thm:search} leaves the selection error $\eps_{\mathrm{sel}}$: even when a good harness has been proposed, the run must recognize it from measured losses. Selection has two stages (Algorithm~\ref{alg:bilevel}). Within a round, the candidate with the lowest training loss is kept, and these candidates together with $h_0$ form the archive $\mathcal V$; at the end, the archived harness with the lowest validation loss is selected. The first stage uses tasks whose trajectories the optimizer has read; the second uses tasks of which it has seen only validation losses and outcome summaries. The lemma charges each stage for its estimation error.

\begin{lemma}[Selection in two stages]
\label{lem:screen}
Suppose that with probability at least $1-\delta$, every training loss is within $a_{\mathrm{train}}$ of its candidate's population loss and every validation loss on $\mathcal V$ is within $a_{\mathrm{val}}$. Then, with $c=\min\{1,2a_{\mathrm{train}}+2a_{\mathrm{val}}\}$, we have $\eps_{\mathrm{sel}}\le c+(1-c)\delta$.
\end{lemma}
\begin{proof}
Let $h_*$ minimize $L$ on $\CC$. On the stated event, if $h_*$ was proposed in a round whose kept candidate is $h^{\mathrm{scr}}$, then
\[
L(h^{\mathrm{scr}})\le\widehat L_{\mathrm{train}}(h^{\mathrm{scr}})+a_{\mathrm{train}}
\le\widehat L_{\mathrm{train}}(h_*)+a_{\mathrm{train}}
\le L(h_*)+2a_{\mathrm{train}},
\]
and if $h_*=h_0$ it is in $\mathcal V$ already; so $\min_{\mathcal V}L\le\min_{\CC}L+2a_{\mathrm{train}}$. The same comparison for validation selection on $\mathcal V$ gives $L(\widehat h)\le\min_{\mathcal V}L+2a_{\mathrm{val}}$. The gap $L(\widehat h)-\min_{\CC}L$ is therefore at most $c$ on the event and at most $1$ off it; take expectations.
\end{proof}
This is the usual comparison between empirical and population minimizers~\citep{hastie2009elements}, applied twice. It also covers the final step of the protocol, which re-evaluates the two archived harnesses with the lowest validation losses and keeps the one with the lower mean: the minimizer of $L$ on $\mathcal V$ is either among the two, and then the winner's mean loss is no larger than its own, or it is not, and then both finalists have first losses no larger than its own. In both cases $L(\widehat h)\le\min_{\mathcal V}L+2a_{\mathrm{val}}$ as long as first losses and mean losses are within $a_{\mathrm{val}}$ of the population loss. The optimizer builds its candidates from the training trajectories, so $a_{\mathrm{train}}$ is a condition on the training comparison rather than a consequence of sampling. In our implementation, a candidate that fails the smoke check is repaired before validation without being screened again (Appendix~\ref{app:algorithm}); such repairs address loading and runtime errors. The lemma covers this case with the repaired version as the candidate in $\CC$ and its training loss measured before the repair. The condition on $a_{\mathrm{train}}$ then compares this measured loss with the population loss of the repaired version. The next subsection bounds $a_{\mathrm{val}}$.

\subsubsection{Validation losses under feedback}
\label{app:selection}
\label{app:evaluation}
The optimizer never executes validation tasks, but their losses and outcome summaries are returned to it and shape later candidates. Each round therefore chooses a harness after looking at the validation set through this feedback. This is adaptive data analysis: how much the validation loss can flatter the chosen harness is governed by how much the feedback reveals, and can be bounded by the number of possible values of the feedback~\citep{dwork2015generalization,blum2015ladder} or by its entropy~\citep{russo2016controlling,xu2017information}. \citet{bertran2026fits} apply the same argument to research agents that receive one bit of validation feedback per submission. We give both forms.

Let $X_1,\ldots,X_m\sim D$ be the validation tasks and, for each harness $h$ and run index $r$, let $Y_i^{(r)}(h)\sim\pi_\theta(\cdot\mid X_i;h)$ be a complete trajectory; given the task, reruns are independent. Collect everything about task $i$ in $Z_i=(X_i,(Y_i^{(r)}(h))_{h,r})$. The $Z_i$ are i.i.d.\ and independent of $U$, the training data and the random seeds that drive the optimizer and every training-side execution. A validation estimate is an average $\widehat\phi=\frac1m\sum_i\phi(Z_i)$ of a per-task statistic $\phi(Z_i)\in[0,1]$ with mean $\bar\phi$. The validation loss of $h$ uses one run per task, $\widehat L_{\mathrm{val}}(h)=\frac1m\sum_i\ell(X_i,Y_i^{(1)}(h))$, and has mean $L(h)$.

Let $F=(F_0,\ldots,F_{T-1})$ be the feedback returned across rounds. We assume that $F$ takes values in a finite set $\mathcal F$, as it does when every summary has bounded length. Our summaries list at most 30 task identifiers and the two most frequent error messages. The number of possible summaries can still be large, and the bounds below are informative only when $\log|\mathcal F|$ is small relative to $m$. Given $U$ and $F_0,\ldots,F_{t-1}$, the run is deterministic up to the validation of round $t$: the validated harness, the statistics that evaluate it, and the optimizer harness $g_t$ are all functions of the feedback so far.

\begin{proposition}[Uniform deviation of validation estimates]
\label{prop:selection}
Let each validated harness receive at most $R$ estimates. For $0<\delta<1$, with probability at least $1-\delta$ every estimate the run computes satisfies
\[
|\widehat\phi-\bar\phi|\le\eps_m=\sqrt{\frac{\log|\mathcal F|+\log(2R(T+1)/\delta)}{2m}},
\]
and so does every average of such estimates.
\end{proposition}
\begin{proof}
Condition on $U=u$; the $Z_i$ keep their law. The harness of round $t\in\{0,\ldots,T\}$, where round $0$ validates $h_0$, and each of its $R$ statistics are fixed once $F_0,\ldots,F_{t-1}$ are, and this prefix takes at most $|\mathcal F|$ values. So at most $R(T+1)|\mathcal F|$ statistics, each fixed before the $Z_i$ are drawn, can ever be computed. Hoeffding's inequality~\citep{hoeffding1963probability} gives $\Prb(|\widehat\phi-\bar\phi|>\eps)\le2e^{-2m\eps^2}$ for each, and a union bound at $\eps=\eps_m$ leaves failure probability $\delta$. Averages inherit the bound by the triangle inequality; integrate over $u$.
\end{proof}

\begin{proposition}[Expected optimism]
\label{prop:entropy}
Let $\widehat h$ be an archived harness chosen by any rule. Then, with entropies in nats,
\[
\E\bigl[L(\widehat h)-\widehat L_{\mathrm{val}}(\widehat h)\bigr]
\le\sqrt{\frac{H(F\mid U)+\log(T+1)}{2m}}
\le\sqrt{\frac{\log|\mathcal F|+\log(T+1)}{2m}}.
\]
\end{proposition}
\begin{proof}
Condition on $U=u$. The archive is a function of $F$ and $\widehat h$ is its entry at some index $J\in\{0,\ldots,T\}$, so $H(\widehat h\mid U=u)\le H(F\mid U=u)+\log(T+1)$. The loss $\ell(X_i,Y_i^{(1)}(h))\in[0,1]$ is $\tfrac12$-sub-Gaussian for every $h$, and $\widehat h$ is a function of $Z_{1:m}$. Theorem~1 of \citet{xu2017information} with $\sigma=\tfrac12$ bounds the expected gap by $\sqrt{I(Z_{1:m};\widehat h\mid U=u)/(2m)}$, and $I(Z_{1:m};\widehat h\mid U=u)\le H(\widehat h\mid U=u)$. Average over $u$ with Jensen's inequality, and use $H(F\mid U)\le\log|\mathcal F|$.
\end{proof}

Both bounds depend on the feedback, not on the harness. The size and content of what the optimizer writes do not enter them, and the optimizer harness $g_t$ is itself a function of $U$ and the feedback so far, so both loops are covered by the same feedback. What matters is $\log|\mathcal F|$. With no feedback it is $0$ and only the choice among $T+1$ archived harnesses remains; a validation loss per round contributes at most $\log(m+1)$; an improved-or-not bit contributes $\log2$; a list of which tasks changed outcome contributes up to $m\log3$; and returning validation trajectories would make $|\mathcal F|$ far larger. This is the trade-off behind the protocol of Section~\ref{sec:splits}. Feedback can raise $r_i$ and lower $e_i$ in Theorem~\ref{thm:search} by showing which edits held up on unseen tasks, and it enlarges $a_{\mathrm{val}}$ through $\log|\mathcal F|$; restricting its possible values lowers this cost.

The protocol of Appendix~\ref{app:algorithm} also reruns a validation task whose outcome differs from the parent's and keeps the change only if it reproduces. Such a confirmed estimate is an average of per-task statistics and is covered by Proposition~\ref{prop:selection}, but its mean is not exactly $L(h)$. On a task where $h$ fails with probability $p$, let $a$ be its first outcome, $b$ the parent's, and $r$ the rerun; the confirmed outcome equals $a$ when $a=b$ and $r$ otherwise, so $\E[c-a\mid x]=p\,\Prb(a=0,b=1)-(1-p)\,\Prb(a=1,b=0)$, and both terms lie in $[0,p(1-p)]$. The mean of a confirmed estimate is therefore within $v(h)=\E_{x\sim D}[p_h(x)(1-p_h(x))]\le\tfrac14$ of $L(h)$, with $v(h)=0$ when outcomes are deterministic given the task. In Lemma~\ref{lem:screen} this term adds to $a_{\mathrm{val}}$: with the two estimates per harness that the final step uses, $a_{\mathrm{val}}=\eps_m$ at $R=2$ plus $\max_h v(h)$.

\section{Additional experiment details}
\label{app:details}

\subsection{Data and grading}
\label{app:data}

\textbf{SWE-rebench splits.}
The training and validation pools draw from the monthly leaderboard pool of the benchmark. The test split was frozen separately, before any method development, and an n-gram overlap scan between the training or validation task texts and the test tasks found no contamination. The sealed test pool holds 110 tasks. Two of them fail their own official reference solution under our grading pipeline and are excluded, so in-distribution SWE-rebench accuracies use 108 tasks.

\textbf{Out-of-distribution test set.}
The OOD test set of Table~\ref{tab:equip-open} is the July 2026 release of the SWE-rebench leaderboard, which covers tasks from May 15 to July 1, 2026. It holds 111 tasks in Go (21), Java (21), Python (20), Rust (25), and TypeScript (24). Four Java tasks fail their reference solution in our environment and are excluded, so OOD accuracies use 107 tasks: Go 21, Java 17, Python 20, Rust 25, and TypeScript 24. On the 87 tasks outside Python, blank scores $25.3\%$, and \method{} with self-evolution scores $35.6\%$ on the Meta-Harness host and $37.9\%$ on the HarnessX host, the two hosts with the largest OOD gains in Table~\ref{tab:equip-open}. Three of the 20 Python tasks come from a repository that also appears in the training and validation sets; the in-distribution test set shares no repository with them.

\textbf{Terminal-Bench splits.}
The Terminal-Bench~\citep{terminalbench2026} pools follow the suite generations, which share no tasks. Training uses 104 Terminal-Bench 1.x tasks, and the test split is the 89 tasks of Terminal-Bench 2.x at a pinned commit, so training never sees a test environment. The validation pool holds 98 tasks (20 from Terminal-Bench 1.x and 78 from Terminal-Bench-Lite), and each run samples 50 of them with a fixed seed.

\textbf{Grading.}
Each task is graded by its own test script, and task environments are frozen as image snapshots, so the checker version cannot drift during a run. Every validation task passed a screening check that its official reference solution passes our grading pipeline. Grader logs are available on training tasks only.

\subsection{Models and budgets}
\label{app:models}

\textbf{Models.}
Sections~\ref{sec:observations} and~\ref{sec:experiments} use Qwen3.8-Flash-Next as the optimizer and Qwen3.8-27B as the executor, served through vLLM; the executor runs with low reasoning effort and greedy decoding. The appendix also reports results with closed-weight models: the small tier of an anonymized model family as the executor and its mid-size tier as the optimizer. We refer to them as the small model and the mid-size model. These models satisfy the two conditions \citet{rethinking2026} identify for informative harness evolution studies: the executor has headroom, and performance depends strongly on the harness. Observation 2 (Figures~\ref{fig:census} and~\ref{fig:valcurves}) also uses four further open-weight executors: Qwen3-Coder-480B, DeepSeek V3.2, Mistral Large 3, and GLM-5. The first two also serve as target executors in the generalization study of Appendix~\ref{app:crossmodel}.

\textbf{Budget.}
Every method runs $T = 6$ evolution rounds on SWE-rebench and $T = 3$ on Terminal-Bench, with three candidates generated per round, the same proposal-turn limit, the same sandbox parallelism, and the same harness edit space (executable hooks plus prompts). The only axis that varies between methods is what the optimizer is given: the evidence interface, the verification tools, and the right to edit its own harness.

\textbf{Simple verification budget.}
In Table~\ref{tab:observations}, simple verification executes one training task under an optional draft edit in an isolated copy of the executor harness. We call each such execution, and each call to a \method{} verification tool, a \emph{probe}. Each call returns the test verdict, the trajectory path, and the remaining execution budget. The configured allowance is eight calls per candidate, scaled by the three candidates to a shared pool of 24 calls per round. The three candidates and the subsequent optimizer self-update draw from this same pool. A passing probe does not automatically accept an edit; the optimizer decides whether to revise, probe again, or submit, after which the shared screening and validation procedure applies.

\textbf{Candidate search and verification costs.}
Each candidate has at most 60 optimizer turns. The candidates are compared on a shared subset of up to 12 training tasks within that round; the same comparison procedure and validation-selection rule apply to baselines and \method{}. A failed generation can leave fewer than three candidates.

For variants with verification, the configured allowance is eight experiment units per candidate, pooled into a shared limit of 24 units per round. This is one common pool, not three independently enforced limits: calls to the three verification tools and, for self-evolving optimizers, to simple verification, including calls during the optimizer self-update, draw from it. In the four self-evolving runs of Table~\ref{tab:equip-open}, simple verification was called 21 times, against 140 calls to the three verification tools. A \texttt{verification} call runs at most three target tasks and costs one unit per target task; a \texttt{replay} or \texttt{perturbation} call costs one unit. Units count experiments, not physical executions: a successful \texttt{verification} call includes a confirmation rerun, and a \texttt{perturbation} call can require both a baseline and a modified replay. All these calls use $\Str$ only.

Baselines without verification receive no probe allowance, and these extra executions are not offset by reducing other runs. Automatic attribution has a separate time budget; optimizer self-updates also require additional model calls. Thus the comparisons match candidate count, proposal-turn limits, and the evaluation protocol, but not total execution or token cost; Appendix~\ref{app:cost} reports the cost of each run. Fixed and self-evolving variants with verification share the same probe allowance.

\subsection{Cost of evolution}
\label{app:cost}

\begin{table}[t]
\caption{\textbf{Cost of each evolution run of Table~\ref{tab:equip-open}}, counted from the run logs. \emph{Agent runs}: runs of the executor on one task, with model calls, excluding test evaluation; \emph{Verification}: the agent runs of the verification tools and simple verification, including their confirmation reruns; \emph{Units}: the verification budget spent; \emph{Replays}: re-executions of recorded commands without model calls; \emph{Tokens}: executor tokens in the training, screening, and validation runs (millions); \emph{Opt.\ calls}: optimizer model turns while generating candidates and during self-updates; \emph{Hours}: wall-clock time from the start of round 1 to the end of harness selection.}
\label{tab:cost}
\vspace{7pt}
\centering
\small
\setlength{\tabcolsep}{3.5pt}
\renewcommand{\arraystretch}{1.05}
\begin{tabular}{l ccccccc}
\toprule
Method & Agent runs & Verification & Units & Replays & Tokens & Opt.\ calls & Hours \\
\midrule
Meta-Harness~\citep{metaharness2026} & 1{,}326 & -- & -- & 0 & 550 & 1{,}030 & 21.6 \\
\quad $+$ \method{} (w/o self-evolve) & 1{,}374 & 39 & 35 & 12 & 567 & 1{,}042 & 32.6 \\
\quad $+$ \method{} (w/ self-evolve) & 1{,}453 & 95 & 67 & 20 & 634 & 1{,}126 & 38.0 \\
\addlinespace[1pt]
\arrayrulecolor{black!50}\hdashline[1.8pt/2.2pt]
\arrayrulecolor{black}\addlinespace[2pt]
AHE~\citep{ahe2026} & 1{,}150 & -- & -- & 0 & 378 & 1{,}035 & 15.3 \\
\quad $+$ \method{} (w/o self-evolve) & 1{,}384 & 27 & 27 & 15 & 731 & 1{,}032 & 28.5 \\
\quad $+$ \method{} (w/ self-evolve) & 1{,}399 & 44 & 43 & 19 & 562 & 715 & 23.8 \\
\addlinespace[1pt]
\arrayrulecolor{black!50}\hdashline[1.8pt/2.2pt]
\arrayrulecolor{black}\addlinespace[2pt]
Self-Harness~\citep{selfharness2026} & 1{,}304 & -- & -- & 0 & 504 & 1{,}058 & 21.5 \\
\quad $+$ \method{} (w/o self-evolve) & 1{,}404 & 33 & 33 & 16 & 628 & 1{,}042 & 28.3 \\
\quad $+$ \method{} (w/ self-evolve) & 1{,}467 & 104 & 86 & 18 & 516 & 931 & 32.9 \\
\addlinespace[1pt]
\arrayrulecolor{black!50}\hdashline[1.8pt/2.2pt]
\arrayrulecolor{black}\addlinespace[2pt]
HarnessX~\citep{harnessx2026} & 1{,}035 & -- & -- & 0 & 443 & 1{,}030 & 19.0 \\
\quad $+$ \method{} (w/o self-evolve) & 1{,}388 & 39 & 36 & 11 & 494 & 1{,}051 & 29.9 \\
\quad $+$ \method{} (w/ self-evolve) & 1{,}441 & 70 & 73 & 37 & 700 & 900 & 47.0 \\
\bottomrule
\end{tabular}
\end{table}

Table~\ref{tab:cost} reports what each run executed. An agent run is one run of the executor on one task. Agent runs comprise the training, screening, and validation runs of every round, the re-evaluation of the two finalists, the confirmation reruns of validation changes, the smoke checks, and the agent runs of the verification tools. Replays re-execute recorded commands in a fresh container without calling a model; trace minimization and the \texttt{replay} tool use one replay, and a \texttt{perturbation} call uses two, one with the original step and one with the changed step.

Units count verification experiments rather than executions: a \texttt{verification} call costs one unit per target task, but each target that passes also gets a confirmation run. Optimizer tokens were not logged, so the table counts optimizer model turns instead. Test evaluation is the same for every method and adds three runs of the 108 in-distribution and 107 out-of-distribution tasks for each evaluated harness.

Training, screening, and validation runs account for 91 to 98\% of the agent runs of every evolution run. The verification tools add 27 to 104 agent runs, 2 to 7\% of a run's agent runs, and at most 37 replays. The larger gaps in agent runs between a baseline and its \method{} versions come from rounds in which none of the baseline's candidates could be applied, so the round ran no screening or validation (AHE and HarnessX). The self-evolving optimizer on the AHE host ends each candidate after a median of 26 turns instead of 58, so it makes fewer optimizer calls than the baseline despite its self-updates.

\method{} runs take 1.3 to 2.5 times as long as their baselines in wall-clock time. Waiting for verification runs while candidates are generated accounts for 1.0 to 4.7 hours of this. Most of the rest is screening and validation, which take longer when an evolved harness makes each executor run longer and when a baseline skipped rounds; the load on the shared executor servers also differed between runs.

\subsection{Counting draft evaluations}
\label{app:mechanism}

The top panel of Figure~\ref{fig:mechanism} and the draft counts of Section~\ref{sec:mechanism} follow these rules. In the evolution runs of Tables~\ref{tab:observations}--\ref{tab:ablation-open}, the \texttt{verification} tool made 170 draft evaluations. Each run of the tool counts as one evaluation, so a draft tested twice counts twice. The tool reports a tested task as passed only after it passes twice. For each passed task, we check whether the harness being edited passed it in its own training run. This sorts the 170 evaluations into four groups:
\begin{itemize}[leftmargin=1.4em,itemsep=0pt,topsep=2pt]
\item in 17, the draft fixed a task that was failing;
\item in 30, it passed only on tasks that already passed, added by the optimizer as regression checks;
\item in 112, it passed none of its tasks;
\item 11 were inconclusive: 10 inconclusive verdicts and one pass whose task is missing from the log.
\end{itemize}
Some drafts in the first two groups became the round's harness unchanged. Across these drafts, all 5 tasks they fixed and 11 of the 12 tasks they used as regression checks passed again in the next training run. When we report how often optimizers revised a draft that passed none of its tasks, a draft counts as revised if the final submission of its candidate differs from it.

\begin{table}[t]
\caption{\textbf{The four baselines as implemented in their original papers}, from the respective publications. In our experiments (Section~\ref{sec:setup}) all methods are reimplemented under one protocol: same executor, same edit space, same candidate count and proposal-turn limit, and the same optimizer model for every method; only the evidence interface of each method is preserved.}
\label{tab:baselineimpl}
\vspace{7pt}
\centering
\footnotesize
\setlength{\tabcolsep}{3pt}
\renewcommand{\arraystretch}{1.25}
\begin{tabular}{p{1.8cm} p{3.2cm} p{2.0cm} p{2.8cm} p{2.5cm}}
\toprule
& Optimizer implementation & Optimizer model & Evidence interface & Edit check \\
\midrule
Meta-Harness & off-the-shelf coding agent (Claude Code) plus a small skill file & Claude Opus 4.6 & prior harnesses, scores, and raw traces, read through the filesystem & none before submission; the next round scores the edit \\
AHE & hand-built pipeline (Evolve Agent in the authors' NexAU framework) & GPT-5.4 & evidence corpus distilled from trajectories by an LLM debugger & prediction logged at edit time, checked against the next round's outcomes \\
Self-Harness & hand-built propose, evaluate, accept loop on the DeepAgents SDK & same model as the executor & deterministic clustering of failed traces into weakness patterns & regression test gate run by the outer loop \\
HarnessX & hand-built four-stage workflow (Digester, Planner, Evolver, Critic) & one meta agent LLM (not stated in the main text) & on-demand trajectory summaries from the digester stage & deterministic checks gate shipping \\
\midrule
\method{} (ours) & agentic loop whose workspace and prompts are themselves optimized & Qwen3.8-Flash-Next; mid-size model (Appendix~\ref{app:models}) & raw traces plus attribution, trace minimization, and a training audit & the \texttt{verification} tool executes the edit before submission, called by the optimizer itself \\
\bottomrule
\end{tabular}
\end{table}

\section{Baseline reimplementation details}
\label{app:baselines}

Table~\ref{tab:baselineimpl} compares how the four published harness optimizers are implemented in their original papers. All four keep the optimizer fixed while it evolves the executor harness; they differ in how the optimizer is built, which model powers it, and how an edit is checked. \method{} differs on the two axes this paper studies: the optimizer harness is itself optimized, and the execution check is a tool the optimizer calls before submitting an edit, not an outer-loop gate applied afterwards.

\textbf{Evidence interfaces.}
The four methods differ in the evidence interface through which the optimizer digests failures: Meta-Harness reads raw traces directly, AHE distills them into per-task root-cause reports, Self-Harness mines model-specific weakness patterns, and HarnessX summarizes trajectories on demand.

\textbf{Evaluation protocols of the original papers.}
The original papers use different evaluation protocols. Meta-Harness uses separate test sets in several domains and reports its Terminal-Bench 2 results on the tasks used for search~\citep{metaharness2026}; AHE and HarnessX report results on the task set used for evolution~\citep{ahe2026,harnessx2026}; Self-Harness hides held-out traces from the proposer and uses held-out scores to accept edits~\citep{selfharness2026}. We therefore evaluate all methods under the protocol of Section~\ref{sec:splits}.

Our reimplementations preserve each method's evidence interface, the column that distinguishes the four systems in Table~\ref{tab:baselineimpl}. Everything else is unified for the controlled comparison: all four run inside our engine with the same executor and harness edit space (executable hooks plus prompts), the same optimizer model, the same candidate count and proposal-turn limit, the same round structure, the same harness selection rule on the validation split, and the same evaluation protocol (Section~\ref{sec:setup}). The original systems use different executors, edit spaces, optimizer models, acceptance rules, and evaluation protocols.

Section~\ref{sec:splits} explains how our shared data splits separate optimization feedback from final test performance. Test-time scaling methods are not comparable baselines under this protocol, because they produce per-instance adaptations rather than a reusable harness.

\section{Additional main results}
\label{app:additional}
\label{app:closed}

This section repeats the comparison of Section~\ref{sec:main-results} with the small model as the executor and the mid-size model as the optimizer (Appendix~\ref{app:models}). It covers SWE-rebench and Terminal-Bench, and repeats the evolution three times for the four baselines and for \method{} on the HarnessX host.

\begin{table}[t]
\caption{\textbf{Each baseline equipped with \method{}, small and mid-size models} (avg@3 accuracy, \%, $\pm$SEM over three test evaluations; one evolution run per entry). ``$+$ \method{} (w/o self-evolve)'' adds the verification tools to the baseline above; ``w/ self-evolve'' also lets the optimizer edit its own harness. $r^{\star}$: the round selected on validation; \emph{Val-selected}: the harness of that round; \emph{Last}: the final-round harness. \textbf{Bold}: higher mean than the baseline of its block. \underline{Underline}: w/ self-evolve better than w/o.}
\label{tab:equip}
\vspace{7pt}
\centering
\small
\setlength{\tabcolsep}{4pt}
\renewcommand{\arraystretch}{1.05}
\begin{tabular}{l c cc c cc}
\toprule
& \multicolumn{3}{c}{\textbf{SWE-rebench}} & \multicolumn{3}{c}{\textbf{Terminal-Bench}} \\
\cmidrule(lr){2-4} \cmidrule(lr){5-7}
Method & $r^{\star}$ & Val-selected & Last & $r^{\star}$ & Val-selected & Last \\
\midrule
Blank (initial harness $h_0$) & -- & $28.70_{\pm0.54}$ & $28.70_{\pm0.54}$ & -- & $37.08_{\pm1.12}$ & $37.08_{\pm1.12}$ \\
\midrule
Meta-Harness~\citep{metaharness2026} & 6 & $30.25_{\pm0.31}$ & $30.25_{\pm0.31}$ & 3 & $37.08_{\pm1.94}$ & $37.08_{\pm1.94}$ \\
\quad $+$ \method{} (w/o self-evolve) & 2 & $\mathbf{31.18}_{\pm0.31}$ & $\mathbf{33.03}_{\pm0.81}$ & 1 & $\mathbf{39.33}_{\pm1.72}$ & $\mathbf{38.96}_{\pm1.35}$ \\
\quad $+$ \method{} (w/ self-evolve) & 5 & $\underline{\mathbf{32.10}}_{\pm2.94}$ & $28.40_{\pm1.34}$ & 3 & $\mathbf{37.83}_{\pm1.63}$ & $\mathbf{37.83}_{\pm1.63}$ \\
\addlinespace[1pt]
\arrayrulecolor{black!50}\hdashline[1.8pt/2.2pt]
\arrayrulecolor{black}\addlinespace[2pt]
AHE~\citep{ahe2026} & 2 & $29.32_{\pm0.31}$ & $28.70_{\pm1.60}$ & 3 & $38.57_{\pm4.79}$ & $38.57_{\pm4.79}$ \\
\quad $+$ \method{} (w/o self-evolve) & 4 & $\mathbf{32.71}_{\pm0.31}$ & $\mathbf{32.41}_{\pm0.54}$ & 3 & $38.57_{\pm1.35}$ & $38.57_{\pm1.35}$ \\
\quad $+$ \method{} (w/ self-evolve) & 5 & $\underline{\mathbf{34.88}}_{\pm1.63}$ & $\underline{\mathbf{33.64}}_{\pm1.23}$ & 1 & $\underline{\mathbf{39.33}}_{\pm2.25}$ & $\underline{\mathbf{38.96}}_{\pm2.28}$ \\
\addlinespace[1pt]
\arrayrulecolor{black!50}\hdashline[1.8pt/2.2pt]
\arrayrulecolor{black}\addlinespace[2pt]
Self-Harness~\citep{selfharness2026} & 4 & $31.18_{\pm2.22}$ & $27.47_{\pm0.81}$ & 3 & $38.20_{\pm0.65}$ & $38.20_{\pm0.65}$ \\
\quad $+$ \method{} (w/o self-evolve) & 6 & $\mathbf{32.71}_{\pm1.63}$ & $\mathbf{32.71}_{\pm1.63}$ & 2 & $\mathbf{40.08}_{\pm0.75}$ & $38.20_{\pm1.12}$ \\
\quad $+$ \method{} (w/ self-evolve) & 1 & $31.18_{\pm0.81}$ & $\mathbf{29.01}_{\pm1.63}$ & 3 & $\underline{\mathbf{40.45}}_{\pm1.72}$ & $\underline{\mathbf{40.45}}_{\pm1.72}$ \\
\addlinespace[1pt]
\arrayrulecolor{black!50}\hdashline[1.8pt/2.2pt]
\arrayrulecolor{black}\addlinespace[2pt]
HarnessX~\citep{harnessx2026} & 5 & $30.86_{\pm0.62}$ & $30.56_{\pm0.53}$ & 2 & $37.08_{\pm0.65}$ & $36.71_{\pm0.99}$ \\
\quad $+$ \method{} (w/o self-evolve) & 5 & $\mathbf{32.41}_{\pm1.06}$ & $29.63_{\pm0.00}$ & 3 & $\mathbf{38.57}_{\pm0.37}$ & $\mathbf{38.57}_{\pm0.37}$ \\
\quad $+$ \method{} (w/ self-evolve) & 5 & $\underline{\mathbf{33.02}}_{\pm1.11}$ & $\underline{\mathbf{32.41}}_{\pm0.53}$ & 3 & $\underline{\mathbf{41.20}}_{\pm1.63}$ & $\underline{\mathbf{41.20}}_{\pm1.63}$ \\
\bottomrule
\end{tabular}
\end{table}

Table~\ref{tab:equip} reports one evolution run per entry, Table~\ref{tab:main} the mean of three runs per method, and Table~\ref{tab:replicates} every run. Every run starts from $h_0$; at $r^{\star}=0$ an entry re-evaluates $h_0$.

\textbf{The verification tools help every baseline.} On SWE-rebench, adding them improves all four baselines at the selected round, and every equipped baseline reaches or exceeds the strongest unequipped one. On Terminal-Bench, the equipped baselines match or improve on their unequipped versions, with margins close to the evaluation noise.

\textbf{\method{} is the strongest configuration on both benchmarks.} Its three-run mean is the highest in both views on both benchmarks, and at the selected round each of its runs scores above the mean of every baseline. The three-run study uses the HarnessX host; the AHE-hosted run of Table~\ref{tab:equip} scores higher on SWE-rebench ($34.88$) but is a single run. Terminal-Bench is less sensitive to the harness in this setting: most methods stay within about two points of blank, in line with \citet{rethinking2026}.

\textbf{The final round is less stable without verification tools.} On SWE-rebench, three of the four baselines lose accuracy from the selected to the final round, and a single run can lose several points. With verification tools, the two views usually stay within two points of each other, although some runs lose more.

\begin{table}[t]
\caption{\textbf{Three-run means with the small and mid-size models} (avg@3 accuracy, \%, $\pm$SEM across three evolution runs, except for Blank, where it is over three test evaluations of $h_0$; every run is listed in Table~\ref{tab:replicates}). \emph{Val-selected}: the harness of the round selected on validation; \emph{Last}: the final-round harness. \method{} (\textbf{bold}) uses the HarnessX host.}
\label{tab:main}
\vspace{7pt}
\centering
\small
\setlength{\tabcolsep}{4pt}
\renewcommand{\arraystretch}{1.05}
\begin{tabular}{l cc cc}
\toprule
& \multicolumn{2}{c}{\textbf{SWE-rebench}} & \multicolumn{2}{c}{\textbf{Terminal-Bench}} \\
\cmidrule(lr){2-3} \cmidrule(lr){4-5}
Method & Val-selected & Last & Val-selected & Last \\
\midrule
Blank (initial harness $h_0$) & $28.70_{\pm0.54}$ & $28.70_{\pm0.54}$ & $37.08_{\pm1.12}$ & $37.08_{\pm1.12}$ \\
Meta-Harness~\citep{metaharness2026} & $30.56_{\pm0.99}$ & $28.71_{\pm2.70}$ & $36.95_{\pm0.33}$ & $38.33_{\pm1.07}$ \\
AHE~\citep{ahe2026} & $29.94_{\pm0.94}$ & $29.32_{\pm0.94}$ & $37.45_{\pm0.57}$ & $37.45_{\pm0.57}$ \\
Self-Harness~\citep{selfharness2026} & $29.32_{\pm0.99}$ & $27.78_{\pm0.47}$ & $37.83_{\pm0.37}$ & $39.33_{\pm0.65}$ \\
HarnessX~\citep{harnessx2026} & $28.39_{\pm1.28}$ & $30.15_{\pm0.91}$ & $36.96_{\pm0.54}$ & $37.21_{\pm0.33}$ \\
\textbf{\method{} (ours)} & $\mathbf{32.82}_{\pm0.20}$ & $\mathbf{31.79}_{\pm0.62}$ & $\mathbf{40.57}_{\pm0.45}$ & $\mathbf{40.20}_{\pm0.82}$ \\
\bottomrule
\end{tabular}
\end{table}

The \underline{underlined} run of each method in Table~\ref{tab:replicates} is the run shown in Table~\ref{tab:equip}; the generalization study of Appendix~\ref{app:crossmodel} starts from its harness.

\begin{figure}[t]
\centering
\includegraphics[width=\textwidth]{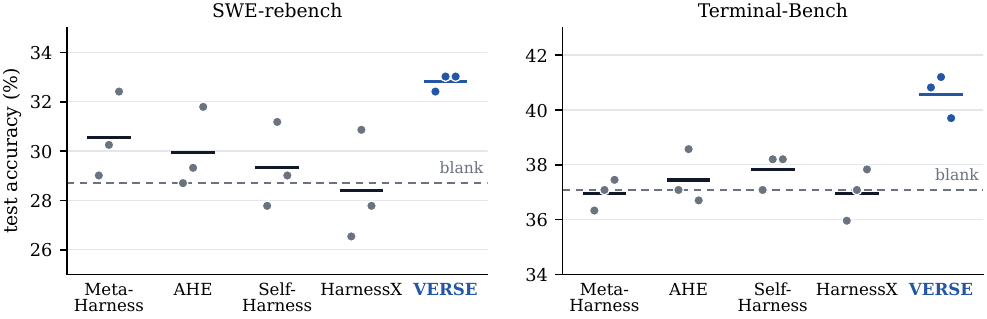}
\par\vspace{-6pt}
\caption{\textbf{The evolution runs behind Table~\ref{tab:main}.} Validation-selected avg@3 test accuracy (\%) of each of the three evolution runs per method (dots; values in Table~\ref{tab:replicates}), with the three-run mean (bar) and blank (dashed line).}
\label{fig:replicate-runs}
\end{figure}

Figure~\ref{fig:replicate-runs} shows the three runs of each method in Table~\ref{tab:main}. On Terminal-Bench, the lowest \method{} run scores above every baseline run.

\begin{table}[t]
\caption{\textbf{Every evolution run behind Table~\ref{tab:main}} (avg@3 accuracy, \%, $\pm$SEM over three test evaluations). Each method is evolved three times from scratch, one run per row; \emph{avg} rows give the mean and the SEM across runs. $r^{\star}$: the round selected on validation; \emph{Val-selected}: the harness of that round; \emph{Last}: the final-round harness. At $r^{\star}=0$ the entry re-evaluates $h_0$. \underline{Underline}: the run shown in Table~\ref{tab:equip}, from which the generalization study starts.}
\label{tab:replicates}
\vspace{7pt}
\centering
\small
\setlength{\tabcolsep}{4pt}
\renewcommand{\arraystretch}{1.12}
\begin{tabular}{l c cc c cc}
\toprule
& \multicolumn{3}{c}{\textbf{SWE-rebench}} & \multicolumn{3}{c}{\textbf{Terminal-Bench}} \\
\cmidrule(lr){2-4} \cmidrule(lr){5-7}
Method & $r^{\star}$ & Val-selected & Last & $r^{\star}$ & Val-selected & Last \\
\midrule
Blank (initial harness $h_0$) & -- & $28.70_{\pm0.54}$ & $28.70_{\pm0.54}$ & -- & $37.08_{\pm1.12}$ & $37.08_{\pm1.12}$ \\
\midrule
\multirow{4}{*}{\begin{tabular}[c]{@{}l@{}}Meta-Harness\\ \citep{metaharness2026}\end{tabular}} & 3 & $29.01_{\pm1.11}$ & $23.46_{\pm0.31}$ & 1 & $36.33_{\pm0.99}$ & $40.45_{\pm1.12}$ \\
& \underline{6} & $\underline{30.25}_{\pm0.31}$ & $\underline{30.25}_{\pm0.31}$ & \underline{3} & $\underline{37.08}_{\pm1.94}$ & $\underline{37.08}_{\pm1.94}$ \\
& 6 & $32.41_{\pm1.06}$ & $32.41_{\pm1.06}$ & 3 & $37.45_{\pm0.37}$ & $37.45_{\pm0.37}$ \\
\addlinespace[1pt]
\arrayrulecolor{black!50}\cdashline{2-7}[1.8pt/2.2pt]
\arrayrulecolor{black}\addlinespace[1.5pt]
& \emph{avg} & $30.56_{\pm0.99}$ & $28.71_{\pm2.70}$ & & $36.95_{\pm0.33}$ & $38.33_{\pm1.07}$ \\
\midrule
\multirow{4}{*}{\begin{tabular}[c]{@{}l@{}}AHE\\ \citep{ahe2026}\end{tabular}} & 3 & $28.70_{\pm0.53}$ & $28.09_{\pm0.31}$ & 3 & $37.08_{\pm2.34}$ & $37.08_{\pm2.34}$ \\
& \underline{2} & $\underline{29.32}_{\pm0.31}$ & $\underline{28.70}_{\pm1.60}$ & \underline{3} & $\underline{38.57}_{\pm4.79}$ & $\underline{38.57}_{\pm4.79}$ \\
& 0 & $31.79_{\pm1.11}$ & $31.17_{\pm1.72}$ & 3 & $36.70_{\pm0.75}$ & $36.70_{\pm0.75}$ \\
\addlinespace[1pt]
\arrayrulecolor{black!50}\cdashline{2-7}[1.8pt/2.2pt]
\arrayrulecolor{black}\addlinespace[1.5pt]
& \emph{avg} & $29.94_{\pm0.94}$ & $29.32_{\pm0.94}$ & & $37.45_{\pm0.57}$ & $37.45_{\pm0.57}$ \\
\midrule
\multirow{4}{*}{\begin{tabular}[c]{@{}l@{}}Self-Harness\\ \citep{selfharness2026}\end{tabular}} & 3 & $27.78_{\pm0.93}$ & $27.16_{\pm2.64}$ & 2 & $37.08_{\pm1.12}$ & $39.33_{\pm1.30}$ \\
& \underline{4} & $\underline{31.18}_{\pm2.22}$ & $\underline{27.47}_{\pm0.81}$ & \underline{3} & $\underline{38.20}_{\pm0.65}$ & $\underline{38.20}_{\pm0.65}$ \\
& 4 & $29.01_{\pm1.55}$ & $28.70_{\pm1.06}$ & 2 & $38.20_{\pm0.65}$ & $40.45_{\pm0.65}$ \\
\addlinespace[1pt]
\arrayrulecolor{black!50}\cdashline{2-7}[1.8pt/2.2pt]
\arrayrulecolor{black}\addlinespace[1.5pt]
& \emph{avg} & $29.32_{\pm0.99}$ & $27.78_{\pm0.47}$ & & $37.83_{\pm0.37}$ & $39.33_{\pm0.65}$ \\
\midrule
\multirow{4}{*}{\begin{tabular}[c]{@{}l@{}}HarnessX\\ \citep{harnessx2026}\end{tabular}} & 5 & $26.54_{\pm0.82}$ & $28.40_{\pm1.54}$ & 2 & $35.96_{\pm1.72}$ & $37.08_{\pm1.12}$ \\
& \underline{5} & $\underline{30.86}_{\pm0.62}$ & $\underline{30.56}_{\pm0.53}$ & \underline{2} & $\underline{37.08}_{\pm0.65}$ & $\underline{36.71}_{\pm0.99}$ \\
& 2 & $27.78_{\pm0.53}$ & $31.48_{\pm0.53}$ & 3 & $37.83_{\pm0.75}$ & $37.83_{\pm0.75}$ \\
\addlinespace[1pt]
\arrayrulecolor{black!50}\cdashline{2-7}[1.8pt/2.2pt]
\arrayrulecolor{black}\addlinespace[1.5pt]
& \emph{avg} & $28.39_{\pm1.28}$ & $30.15_{\pm0.91}$ & & $36.96_{\pm0.54}$ & $37.21_{\pm0.33}$ \\
\midrule
\multirow{4}{*}{\method{} (ours)} & 6 & $32.41_{\pm1.06}$ & $32.41_{\pm1.06}$ & 3 & $40.82_{\pm2.62}$ & $40.82_{\pm2.62}$ \\
& \underline{5} & $\underline{33.02}_{\pm1.11}$ & $\underline{32.41}_{\pm0.53}$ & \underline{3} & $\underline{41.20}_{\pm1.63}$ & $\underline{41.20}_{\pm1.63}$ \\
& 3 & $33.02_{\pm0.62}$ & $30.56_{\pm0.53}$ & 1 & $39.70_{\pm1.63}$ & $38.58_{\pm0.99}$ \\
\addlinespace[1pt]
\arrayrulecolor{black!50}\cdashline{2-7}[1.8pt/2.2pt]
\arrayrulecolor{black}\addlinespace[1.5pt]
& \emph{avg} & $32.82_{\pm0.20}$ & $31.79_{\pm0.62}$ & & $40.57_{\pm0.45}$ & $40.20_{\pm0.82}$ \\
\bottomrule
\end{tabular}
\end{table}

\section{Generalization study}
\label{app:crossmodel}

This appendix asks what an evolved harness carries beyond the setting in which it was evolved. We study three shifts, and they separate what transfers from what does not. With the same executor and the same kind of task, the gains of \method{} carry over from Python to newer tasks in four other programming languages (Section~\ref{app:languages}). When the task domain changes, the executor harness stays specific to its source domain, and what carries over is the optimizer harness learned by self-evolution (Section~\ref{app:crossdomain-study}). When the executor model changes, executor harnesses tuned to one model mostly stop helping (Section~\ref{app:crossmodel-study}). Section~\ref{app:languages} uses the harnesses of Table~\ref{tab:equip-open}; the other two sections start from the harnesses of Table~\ref{tab:equip}, evolved with the small and mid-size models. Throughout, every score is compared to blank, the initial harness $h_0$, measured on the same test set with the same executor.

\subsection{Transfer to new programming languages}
\label{app:languages}

\begin{figure}[t]
\centering
\includegraphics[width=\textwidth]{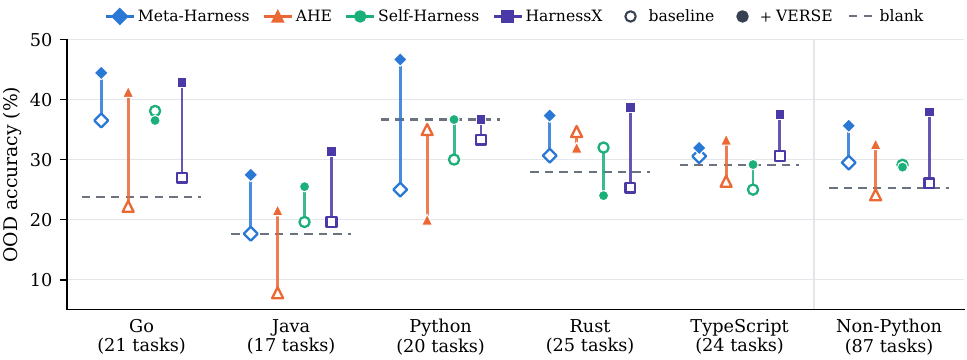}
\par\vspace{-6pt}
\caption{\textbf{OOD accuracy by programming language.} The validation-selected harnesses of Table~\ref{tab:equip-open} on the 107 OOD tasks, split by language (mean over three test evaluations). Each host has one color: the hollow marker is the baseline, the filled marker is the baseline with \method{} and self-evolution, and the line joins the two. All harnesses were evolved on Python tasks.}
\label{fig:ood-languages}
\end{figure}

All training and validation tasks of Section~\ref{sec:splits} are in Python, while 87 of the 107 OOD test tasks of Table~\ref{tab:equip-open} are in Go, Java, Rust, or TypeScript. Each OOD task runs in its own released container image, which provides the toolchain of its language, and is graded by its own test script with the SWE-rebench log parser for that language. Figure~\ref{fig:ood-languages} splits the OOD accuracy of Table~\ref{tab:equip-open} by language. Outside Python, the four baselines range from 1.1 points below blank to 4.2 points above it, while \method{} with self-evolution is 3.4 to 12.6 points above blank. \method{} raises its host in 16 of the 20 pairs of host and language, and in all five languages on the Meta-Harness and HarnessX hosts. Averaged over the four hosts, it raises accuracy in every language: by 10.3 points in Go, 10.3 in Java, 2.3 in Rust, 4.9 in TypeScript, and 4.2 in Python. Self-evolution matters most on these two hosts: without it, the evolved harnesses stay near their baselines outside Python (28.7\% and 26.8\%, against 29.5\% and 26.1\%); with it, they reach 35.6\% and 37.9\%. On the AHE and Self-Harness hosts, the tools alone raise accuracy outside Python (to 35.2\% and 30.3\%, from 24.1\% and 29.1\%).

\subsection{Transfer to a new task domain}
\label{app:crossdomain-study}

Table~\ref{tab:crossdomain} moves the evolved harnesses between SWE-rebench and Terminal-Bench and allows two rounds of evolution in the target domain.

\begin{table}[t]
\caption{\textbf{Cross-domain transfer (SWE $\leftrightarrow$ Terminal-Bench).} Each harness of Table~\ref{tab:equip} is moved to the other domain, evaluated frozen (zero-shot), and evolved for two more rounds there. An evolved optimizer inherits its harness from the final round of the source run; a cold start begins at $h_0$, so its zero-shot entry evaluates $h_0$ and is not bolded. The \method{} rows start from the HarnessX-hosted \method{} harness. Avg@3 accuracy (\%, $\pm$SEM over three test evaluations). \textbf{Bold}: higher mean than the reference, which is the baseline above for ``$+$ \method{}'' rows and the best baseline in the column for self-evolving rows.}
\label{tab:crossdomain}
\vspace{7pt}
\centering
\footnotesize
\setlength{\tabcolsep}{2.5pt}
\renewcommand{\arraystretch}{1.12}
\resizebox{\linewidth}{!}{%
\begin{tabular}{l ccc ccc}
\toprule
& \multicolumn{3}{c}{\textbf{SWE $\to$ Terminal-Bench}} & \multicolumn{3}{c}{\textbf{Terminal-Bench $\to$ SWE}} \\
\cmidrule(lr){2-4} \cmidrule(lr){5-7}
Method & Zero-shot & Round 1 & Round 2 & Zero-shot & Round 1 & Round 2 \\
\midrule
Blank (initial harness $h_0$) & $37.08_{\pm1.12}$ & & & $28.70_{\pm0.54}$ & & \\
\midrule
\multicolumn{7}{l}{\emph{Optimizer fixed; executor harness starts from the source-domain Table~\ref{tab:equip} harness (warm start)}} \\
Meta-Harness & $32.21_{\pm1.88}$ & $36.71_{\pm2.46}$ & $35.20_{\pm0.37}$ & $29.94_{\pm1.11}$ & $29.01_{\pm0.31}$ & $30.86_{\pm0.81}$ \\
\quad $+$ \method{} (w/o self-evolve) & $26.97_{\pm0.65}$ & $\mathbf{39.33}_{\pm0.65}$ & $\mathbf{36.71}_{\pm1.49}$ & $26.85_{\pm2.83}$ & $\mathbf{31.18}_{\pm1.11}$ & $29.63_{\pm0.54}$ \\
\addlinespace[2pt]
AHE & $36.33_{\pm1.35}$ & $38.58_{\pm0.75}$ & $37.08_{\pm0.65}$ & $28.40_{\pm0.31}$ & $32.71_{\pm2.16}$ & $28.70_{\pm1.42}$ \\
\quad $+$ \method{} (w/o self-evolve) & $32.58_{\pm2.34}$ & $34.83_{\pm1.72}$ & $36.71_{\pm1.98}$ & $\mathbf{30.86}_{\pm1.88}$ & $30.56_{\pm2.14}$ & $\mathbf{32.41}_{\pm1.93}$ \\
\addlinespace[2pt]
Self-Harness & $30.71_{\pm2.70}$ & $38.58_{\pm1.98}$ & $35.58_{\pm2.28}$ & $29.94_{\pm0.31}$ & $33.33_{\pm1.93}$ & $31.18_{\pm1.72}$ \\
\quad $+$ \method{} (w/o self-evolve) & $\mathbf{31.46}_{\pm1.29}$ & $34.83_{\pm1.12}$ & $32.96_{\pm0.37}$ & $\mathbf{33.95}_{\pm0.62}$ & $29.94_{\pm0.62}$ & $28.40_{\pm2.22}$ \\
\addlinespace[2pt]
HarnessX & $28.46_{\pm0.37}$ & $35.58_{\pm1.35}$ & $34.46_{\pm1.49}$ & $28.70_{\pm1.60}$ & $27.47_{\pm2.53}$ & $31.79_{\pm2.64}$ \\
\quad $+$ \method{} (w/o self-evolve) & $\mathbf{34.46}_{\pm2.70}$ & $34.83_{\pm1.12}$ & $34.08_{\pm1.63}$ & $27.78_{\pm0.93}$ & $\mathbf{29.63}_{\pm1.85}$ & $\mathbf{34.88}_{\pm1.88}$ \\
\midrule
\multicolumn{7}{l}{\emph{Optimizer evolved; executor harness starts from the source-domain Table~\ref{tab:equip} harness (warm start)}} \\
Self-evolving optimizer & $30.34_{\pm0.00}$ & $35.58_{\pm1.49}$ & $\mathbf{37.83}_{\pm2.62}$ & $\mathbf{31.18}_{\pm0.62}$ & $30.25_{\pm1.11}$ & $30.25_{\pm0.81}$ \\
\method{} (w/ self-evolve) & $\mathbf{36.70}_{\pm1.63}$ & $\mathbf{40.07}_{\pm1.87}$ & $\mathbf{39.33}_{\pm1.12}$ & $\mathbf{31.48}_{\pm1.85}$ & $32.71_{\pm0.81}$ & $30.56_{\pm1.93}$ \\
\addlinespace[3pt]
\multicolumn{7}{l}{\emph{Optimizer evolved; executor harness restarts from blank (cold start)}} \\
Self-evolving optimizer & $37.08_{\pm1.12}$ & $37.08_{\pm0.65}$ & $\mathbf{41.20}_{\pm0.75}$ & $28.70_{\pm0.54}$ & $28.40_{\pm0.81}$ & $27.47_{\pm1.63}$ \\
\method{} (w/ self-evolve) & $36.33_{\pm0.37}$ & $36.70_{\pm0.75}$ & $\mathbf{39.33}_{\pm1.95}$ & $28.70_{\pm0.54}$ & $30.56_{\pm1.93}$ & $30.86_{\pm0.62}$ \\
\bottomrule
\end{tabular}}
\end{table}

Unlike a change of language, a change of domain changes the task itself: a Terminal-Bench task is graded by a program that checks the final state of its container, not by hidden tests applied to a repository diff. Zero-shot, a transplanted executor harness can fall below the target-domain blank: it encodes its source domain, from tool conventions to the failures it fixes. One or two rounds of adaptation bring it back to about the blank level. Gains beyond blank come from the inherited optimizer harness. From SWE-rebench to Terminal-Bench, \method{} with its inherited optimizer is above every frozen-optimizer entry in all three columns, and with the executor harness restarted from blank, one self-evolving configuration matches the best Terminal-Bench score of Table~\ref{tab:equip} within two rounds. The reverse direction moves less, consistent with how flat Terminal-Bench is. What crosses domains appears to be the optimizer's methodology, its skills, notes, and tools for diagnosing and verifying failures, while the executor harness stays specific to its source domain.

\subsection{Transfer to a new executor model}
\label{app:crossmodel-study}

We hand the evolved harnesses to two open-weight executors: Qwen3-Coder-480B and DeepSeek V3.2. Each harness from Table~\ref{tab:equip} (evolved with the small model) is re-evaluated frozen (zero-shot) and then adapted for two rounds with the new executor under the main protocol.

\begin{table}[t]
\caption{\textbf{Transfer to a new executor on SWE-rebench.} The harnesses of Table~\ref{tab:equip}, evolved with the small model, are handed to Qwen3-Coder-480B and DeepSeek V3.2. Each is evaluated frozen with the new executor (zero-shot) and then evolved for two more rounds with it (Rounds 1 and 2). Evolved optimizers inherit their harness from the final round of the source run; a cold start restarts the executor harness at $h_0$, so its zero-shot entry evaluates $h_0$ and is not bolded. The \method{} rows start from the HarnessX-hosted \method{} harness. Avg@3 accuracy (\%, $\pm$SEM over three test evaluations). \textbf{Bold}: higher mean than the reference, which is the baseline above for ``$+$ \method{}'' rows and the best baseline in the column for self-evolving rows.}
\label{tab:crossmodel}
\vspace{7pt}
\centering
\footnotesize
\setlength{\tabcolsep}{3pt}
\renewcommand{\arraystretch}{1.12}
\resizebox{\linewidth}{!}{%
\begin{tabular}{l ccc ccc}
\toprule
& \multicolumn{3}{c}{\textbf{Qwen3-Coder-480B}} & \multicolumn{3}{c}{\textbf{DeepSeek V3.2}} \\
\cmidrule(lr){2-4} \cmidrule(lr){5-7}
Method & Zero-shot & Round 1 & Round 2 & Zero-shot & Round 1 & Round 2 \\
\midrule
Blank (initial harness $h_0$) & $13.89_{\pm0.00}$ & & & $28.08_{\pm2.16}$ & & \\
\midrule
\multicolumn{7}{l}{\emph{Optimizer fixed; executor harness starts from the Table~\ref{tab:equip} harness (warm start)}} \\
Meta-Harness & $13.58_{\pm0.62}$ & $17.29_{\pm1.11}$ & $17.29_{\pm0.31}$ & $26.85_{\pm1.06}$ & $26.55_{\pm2.22}$ & $22.84_{\pm1.72}$ \\
\quad $+$ \method{} (w/o self-evolve) & $\mathbf{16.36}_{\pm3.13}$ & $\mathbf{20.99}_{\pm0.31}$ & $\mathbf{19.75}_{\pm0.81}$ & $26.55_{\pm1.34}$ & $25.31_{\pm2.03}$ & $\mathbf{23.45}_{\pm2.64}$ \\
\addlinespace[2pt]
AHE & $12.04_{\pm0.54}$ & $18.52_{\pm1.93}$ & $9.88_{\pm1.55}$ & $24.38_{\pm1.23}$ & $25.93_{\pm1.42}$ & $25.93_{\pm0.54}$ \\
\quad $+$ \method{} (w/o self-evolve) & $\mathbf{14.51}_{\pm1.63}$ & $13.27_{\pm1.34}$ & $\mathbf{19.75}_{\pm1.11}$ & $\mathbf{27.16}_{\pm0.81}$ & $25.62_{\pm2.03}$ & $25.62_{\pm1.23}$ \\
\addlinespace[2pt]
Self-Harness & $13.89_{\pm0.54}$ & $12.66_{\pm2.16}$ & $13.89_{\pm0.54}$ & $24.69_{\pm1.34}$ & $24.69_{\pm1.34}$ & $25.93_{\pm1.93}$ \\
\quad $+$ \method{} (w/o self-evolve) & $13.27_{\pm1.11}$ & $11.73_{\pm0.81}$ & $13.58_{\pm2.47}$ & $\mathbf{26.55}_{\pm2.03}$ & $\mathbf{25.00}_{\pm1.06}$ & $25.00_{\pm1.85}$ \\
\addlinespace[2pt]
HarnessX & $16.05_{\pm0.31}$ & $17.90_{\pm0.62}$ & $22.84_{\pm1.72}$ & $24.69_{\pm1.63}$ & $33.33_{\pm0.54}$ & $31.17_{\pm0.82}$ \\
\quad $+$ \method{} (w/o self-evolve) & $\mathbf{20.99}_{\pm0.81}$ & $\mathbf{23.77}_{\pm1.11}$ & $22.53_{\pm0.81}$ & $\mathbf{29.01}_{\pm2.41}$ & $31.48_{\pm0.93}$ & $\mathbf{33.03}_{\pm2.03}$ \\
\midrule
\multicolumn{7}{l}{\emph{Optimizer evolved; executor harness starts from the Table~\ref{tab:equip} harness (warm start)}} \\
Self-evolving optimizer & $12.96_{\pm1.06}$ & $17.29_{\pm0.81}$ & $15.44_{\pm1.55}$ & $26.55_{\pm1.11}$ & $24.69_{\pm2.16}$ & $26.23_{\pm0.62}$ \\
\method{} (w/ self-evolve) & $12.65_{\pm0.82}$ & $16.05_{\pm2.02}$ & $18.83_{\pm0.62}$ & $\mathbf{28.70}_{\pm2.83}$ & $28.09_{\pm0.82}$ & $27.16_{\pm2.41}$ \\
\addlinespace[3pt]
\multicolumn{7}{l}{\emph{Optimizer evolved; executor harness restarts from blank (cold start)}} \\
Self-evolving optimizer & $13.89_{\pm0.00}$ & $13.27_{\pm1.55}$ & $17.90_{\pm1.34}$ & $28.08_{\pm2.16}$ & $26.85_{\pm1.93}$ & $24.38_{\pm1.11}$ \\
\method{} (w/ self-evolve) & $13.89_{\pm0.00}$ & $11.11_{\pm1.41}$ & $12.65_{\pm2.16}$ & $28.08_{\pm2.16}$ & $26.54_{\pm1.63}$ & $23.15_{\pm2.78}$ \\
\bottomrule
\end{tabular}}
\end{table}

\textbf{Zero-shot transfer to a new executor model.}
The zero-shot columns of Table~\ref{tab:crossmodel} freeze the ten SWE harnesses of Table~\ref{tab:equip} and swap the executor. A harness evolved for one executor mostly stops helping when the executor changes: averaged over the ten harnesses, zero-shot transfer is close to zero. Two patterns remain. The equipped version keeps its advantage over its baseline in three of four pairs on each executor, and transfer is larger for the executor that handles basic tool use worse under blank (Qwen). The clear exception is the HarnessX family on Qwen: both variants sit above blank, and HarnessX $+$ \method{} gives the largest zero-shot gain in the table.

To see why the HarnessX $+$ \method{} harness transfers, we compared all 108 Qwen test trajectories under the blank harness with the same tasks under this harness. That harness is a minimal prompt plus an 18KB hook file with three components: a \texttt{str\_replace\_editor} tool whose edit content travels as a tool call argument rather than through the shell, reminders that escalate at fixed turn counts if the agent has not yet edited a source file or run the tests, and fixes for two common environment errors.

Under blank, Qwen never uses an editor tool and writes every patch through shell heredocs; in all 8 tasks that the transplant flips, the blank trajectory dies on heredoc quoting before the patch ever lands. Under the transplant, all 108 trajectories use the editor, all 8 flipped tasks land their edits, and trajectories become shorter. \method{}'s own evolved harness, which encodes behavioral fixes specific to the small model, does not transfer. The contrast suggests a mechanism: \emph{harness components that fix tool use errors common across models, such as shell quoting, transfer; components tuned to one model's behavioral quirks do not}.

\begin{table}[t]
\caption{\textbf{Transfer to a new executor on Terminal-Bench.} The Terminal-Bench harnesses of Table~\ref{tab:equip}, evolved with the small model, are handed to Qwen3-Coder-480B and DeepSeek V3.2. Each is evaluated frozen with the new executor (zero-shot) and then evolved for two more rounds with it (Rounds 1 and 2). Evolved optimizers inherit their harness from the final round of the source run; a cold start restarts the executor harness at $h_0$, so its zero-shot entry evaluates $h_0$ and is not bolded. The \method{} rows start from the HarnessX-hosted \method{} harness. Avg@3 accuracy (\%, $\pm$SEM over three test evaluations). \textbf{Bold}: higher mean than the reference, which is the baseline above for ``$+$ \method{}'' rows and the best baseline in the column for self-evolving rows.}
\label{tab:crossmodeltb}
\vspace{7pt}
\centering
\footnotesize
\setlength{\tabcolsep}{3pt}
\renewcommand{\arraystretch}{1.12}
\resizebox{\linewidth}{!}{%
\begin{tabular}{l ccc ccc}
\toprule
& \multicolumn{3}{c}{\textbf{Qwen3-Coder-480B}} & \multicolumn{3}{c}{\textbf{DeepSeek V3.2}} \\
\cmidrule(lr){2-4} \cmidrule(lr){5-7}
Method & Zero-shot & Round 1 & Round 2 & Zero-shot & Round 1 & Round 2 \\
\midrule
Blank (initial harness $h_0$) & $29.96_{\pm0.99}$ & & & $35.21_{\pm1.35}$ & & \\
\midrule
\multicolumn{7}{l}{\emph{Optimizer fixed; executor harness starts from the Table~\ref{tab:equip} harness (warm start)}} \\
Meta-Harness & $27.34_{\pm2.28}$ & $27.34_{\pm0.75}$ & $27.72_{\pm1.35}$ & $31.84_{\pm2.93}$ & $29.21_{\pm0.65}$ & $33.33_{\pm1.35}$ \\
\quad $+$ \method{} (w/o self-evolve) & $\mathbf{31.09}_{\pm0.99}$ & $\mathbf{28.09}_{\pm1.72}$ & $\mathbf{31.09}_{\pm0.99}$ & $29.59_{\pm0.99}$ & $\mathbf{34.83}_{\pm1.72}$ & $\mathbf{34.46}_{\pm0.37}$ \\
\addlinespace[2pt]
AHE & $26.59_{\pm0.75}$ & $28.84_{\pm1.35}$ & $25.47_{\pm1.35}$ & $37.45_{\pm0.99}$ & $35.96_{\pm0.65}$ & $40.45_{\pm2.34}$ \\
\quad $+$ \method{} (w/o self-evolve) & $25.47_{\pm0.99}$ & $\mathbf{29.21}_{\pm0.65}$ & $\mathbf{27.72}_{\pm2.70}$ & $34.46_{\pm1.98}$ & $34.08_{\pm1.63}$ & $33.71_{\pm1.72}$ \\
\addlinespace[2pt]
Self-Harness & $29.21_{\pm1.72}$ & $28.46_{\pm0.99}$ & $28.46_{\pm0.99}$ & $29.21_{\pm2.83}$ & $34.46_{\pm1.63}$ & $31.09_{\pm3.00}$ \\
\quad $+$ \method{} (w/o self-evolve) & $24.34_{\pm1.98}$ & $25.47_{\pm1.50}$ & $25.84_{\pm0.65}$ & $\mathbf{35.96}_{\pm1.30}$ & $29.21_{\pm1.30}$ & $\mathbf{37.83}_{\pm1.35}$ \\
\addlinespace[2pt]
HarnessX & $30.34_{\pm2.25}$ & $27.34_{\pm0.99}$ & $29.21_{\pm0.65}$ & $33.71_{\pm0.65}$ & $31.84_{\pm1.63}$ & $38.95_{\pm1.87}$ \\
\quad $+$ \method{} (w/o self-evolve) & $26.22_{\pm2.09}$ & $27.34_{\pm2.70}$ & $28.84_{\pm1.35}$ & $\mathbf{35.21}_{\pm0.37}$ & $\mathbf{32.96}_{\pm0.37}$ & $37.08_{\pm1.30}$ \\
\midrule
\multicolumn{7}{l}{\emph{Optimizer evolved; executor harness starts from the Table~\ref{tab:equip} harness (warm start)}} \\
Self-evolving optimizer & $28.84_{\pm0.37}$ & $26.97_{\pm0.65}$ & $\mathbf{31.09}_{\pm0.99}$ & $35.58_{\pm0.99}$ & $35.21_{\pm1.98}$ & $36.70_{\pm1.35}$ \\
\method{} (w/ self-evolve) & $\mathbf{31.84}_{\pm0.99}$ & $\mathbf{30.34}_{\pm0.65}$ & $\mathbf{31.09}_{\pm0.37}$ & $35.58_{\pm1.35}$ & $32.96_{\pm2.93}$ & $32.58_{\pm1.30}$ \\
\addlinespace[3pt]
\multicolumn{7}{l}{\emph{Optimizer evolved; executor harness restarts from blank (cold start)}} \\
Self-evolving optimizer & $29.96_{\pm0.99}$ & $\mathbf{30.34}_{\pm2.59}$ & $29.21_{\pm1.30}$ & $35.21_{\pm1.35}$ & $34.83_{\pm2.25}$ & $35.58_{\pm1.87}$ \\
\method{} (w/ self-evolve) & $29.96_{\pm0.99}$ & $\mathbf{29.96}_{\pm0.99}$ & $\mathbf{30.34}_{\pm0.65}$ & $35.21_{\pm1.35}$ & $35.21_{\pm0.75}$ & $31.09_{\pm1.98}$ \\
\bottomrule
\end{tabular}}
\end{table}

\textbf{The same transfer on Terminal-Bench.}
Table~\ref{tab:crossmodeltb} repeats the executor swap for the Terminal-Bench harnesses of Table~\ref{tab:equip}. Zero-shot, most transplanted harnesses stay near the blank, and adaptation recovers the blank level but rarely exceeds it. There is no Terminal-Bench analogue of the HarnessX jump: the Terminal-Bench executor issues one shell command per turn and never passes edits through shell quoting, so the friction that the transferred editor tool removes on SWE-rebench does not arise. The one consistent signal again favors the inherited optimizer harness: on Qwen, \method{} is the only method above the best baseline in all three columns, and its cold-started row also exceeds the best baseline in both adaptation rounds.

\textbf{Warm-started re-evolution on a new executor.}
The Round 1 and Round 2 columns of Table~\ref{tab:crossmodel} ask whether one evolution run can be reused across executor models. In our runs, the best warm-started configuration outscores the best cold-started one on each executor after two rounds. Adaptation helps most when the starting harness is weak on the new executor and little when it is already strong, as for HarnessX $+$ \method{} on Qwen. This suggests that evolving once and then adapting for a round or two can be a practical way to move to a new executor.

\section{Additional ablations}
\label{app:ablations}

\begin{figure}[t]
\centering
\includegraphics[width=0.58\linewidth]{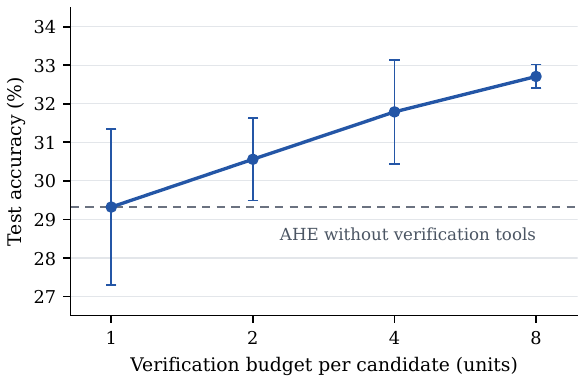}
\par\vspace{-7pt}
\caption{\textbf{The probe budget ablation.} Avg@3 test accuracy (\%, $\pm$SEM over three test evaluations; one evolution run per budget) of AHE $+$ \method{} (w/o self-evolve) as the verification budget per candidate varies; the three candidates of a round share a pool of three times this number of units, and the full configuration allows 8 per candidate (24 per round). The dashed line is AHE without any verification tools.}
\label{fig:budget}
\end{figure}

This appendix uses the small and mid-size models of Appendix~\ref{app:closed} on SWE-rebench. Table~\ref{tab:ablation} collects three ablations. The first repeats the component ablation of Section~\ref{sec:ablation} with these models, again on AHE without self-evolution, the host that gains most from the tools in Table~\ref{tab:equip}. The second keeps this configuration and its probe budget, but the optimizer no longer chooses the probes. The third uses the setting of Observation~1, the self-evolving optimizer with simple verification on Meta-Harness, and restricts what the optimizer may write for itself. Figure~\ref{fig:budget} varies the probe budget. Each configuration in this appendix is a single evolution run, and Table~\ref{tab:replicates} shows that runs of one method can differ by several points, so we read these ablations for their trends.

\begin{table}[t]
\caption{\textbf{Ablations with the small and mid-size models} (SWE-rebench; avg@3 accuracy, \%, $\pm$SEM over three test evaluations; one evolution run per row). Top and middle: AHE $+$ \method{} without self-evolution, with components removed (top) or with the probes scheduled by a rule instead of the optimizer (middle); both change the full configuration in the first row. Bottom: the self-evolving optimizer with simple verification on Meta-Harness, as in Observation~1, restricted in what it may write for itself; the first row is an independent run without the restriction. $r^{\star}$: the round selected on validation; \emph{Val-selected}: the harness of that round; \emph{Last}: the final-round harness.}
\label{tab:ablation}
\vspace{7pt}
\centering
\small
\setlength{\tabcolsep}{4pt}
\renewcommand{\arraystretch}{1.08}
\begin{tabular}{l c cc}
\toprule
Configuration & $r^{\star}$ & Val-selected & Last \\
\midrule
AHE $+$ \method{} (w/o self-evolve), full configuration & 4 & $32.71_{\pm0.31}$ & $32.41_{\pm0.54}$ \\
\multicolumn{4}{l}{\emph{Components removed}} \\
\quad w/o verification during evolution (minimization kept) & 4 & $30.56_{\pm0.53}$ & $32.41_{\pm1.60}$ \\
\quad w/o attribution and minimization & 5 & $30.86_{\pm0.31}$ & $31.17_{\pm1.11}$ \\
\quad w/o training audit only & 1 & $30.25_{\pm0.82}$ & $25.00_{\pm0.53}$ \\
\quad AHE without any \method{} component & 2 & $29.32_{\pm0.31}$ & $28.70_{\pm1.60}$ \\
\midrule
\multicolumn{4}{l}{\emph{Probes not chosen by the optimizer (same probe budget)}} \\
\quad Random schedule & 3 & $30.86_{\pm0.31}$ & $29.01_{\pm2.02}$ \\
\quad Fixed hand-written schedule & 5 & $26.85_{\pm1.60}$ & $30.86_{\pm0.31}$ \\
\midrule
\multicolumn{4}{l}{\emph{What the self-evolving optimizer may write for itself}} \\
Self-evolving $+$ simple verification, unrestricted & 3 & $33.64_{\pm0.81}$ & $27.78_{\pm0.54}$ \\
\quad only its prompts & 2 & $30.25_{\pm1.11}$ & $28.70_{\pm1.60}$ \\
\quad only notes and skills & 2 & $30.56_{\pm1.07}$ & $28.09_{\pm0.82}$ \\
\quad only tools & 1 & $30.25_{\pm0.82}$ & $27.78_{\pm0.53}$ \\
\bottomrule
\end{tabular}
\end{table}

\textbf{The training audit matters most.} Removing attribution and minimization, which also removes the training audit, costs about half of \method{}'s margin over plain AHE. Removing only the training audit, with attribution and minimization still running, costs more, and its final round is the lowest in the table. Attributing individual failures without tracking them across rounds may pull the optimizer toward causes it would otherwise deprioritize.

\textbf{The optimizer choosing its own probes is part of the gain.} At the same probe budget, a random schedule costs accuracy, and a fixed hand-written schedule falls below plain AHE at the selected round: probes the optimizer did not ask for can be worse than none. This suggests that the value of verification lies in the optimizer's own choice of experiments, not in the extra computation alone.

\textbf{No single kind of self-written equipment replaces the full workspace.} Restricting the self-evolving optimizer to only its prompts, only its notes and skills, or only its tools lowers selected-round accuracy by about three points in each case.

\textbf{The gain needs probe budget.} With a budget of one or two units per candidate, most of the gain disappears; from four units on, most of it returns (Figure~\ref{fig:budget}). This is consistent with Appendix~\ref{app:bridge}, where the bound on the repair error falls geometrically with the number of checks, then levels off.

\section{Replay study details}
\label{app:replay}

Each verification tool of Section~\ref{sec:instruments} rests on an empirical premise: attribution and minimization assume that failures compress to a few load-bearing steps, the \texttt{perturbation} tool assumes that single steps carry the outcome, and the \texttt{verification} tool assumes that repairing the diagnosed step can flip it. We test each premise by re-executing recorded traces (Figure~\ref{fig:minrepair}). The measurements below support these premises, and a fourth finding shapes the design: reading a trace identifies the kind of failure far more reliably than the failing step, so localization needs execution.

\subsection{Corpora and replay}

The study uses two corpora. The main corpus holds roughly 300 SWE-rebench task instances~\citep{swerebench2025}, each with a successful and a failed trajectory of an OpenHands agent driven by Qwen3-Coder-480B under one fixed harness. A step is one message of a trajectory: the task prompt, a model turn, or a tool result. Minimization and repair use 299 successful and 298 failed traces; the perturbation experiment injects faults into 297 successful traces. A closed-weight LLM supervisor, a different model from the executor, produces every diagnosis, and every claim is checked by re-executing the trace in the task's Docker environment against the task's own tests.

The second corpus is CodeTraceBench~\citep{codetracer2026}, a public collection of executed trajectories from four agent frameworks (OpenHands, mini-SWE-agent, SWE-agent, and Terminus 2) whose failed runs carry human step-level annotations of the steps on the causal chain to the failure. We use its verified split; the 208 failed trajectories for which a human label and a supervisor diagnosis both exist enter the attribution comparison below.

\textbf{How replay works.} Each replay starts from a fresh container of the task at its original commit. The recorded actions of the kept steps, that is, the agent's commands and file edits, are executed again in order, so every tool output comes from the environment rather than from the recording. After the last kept action, the agent continues on its own until it stops, and its final patch is graded with the task's own tests. The replays inside \method{} stop after the recorded commands and call no model.

A reduced subset is accepted only if its replay fails the same tests as the full trace, or, for a successful trace, still resolves the task; 230 of the 235 accepted subsets of failed traces keep at least one of the agent's own edits, and in the other 5 the failure appears without any agent edit. A repair counts only if a second replay also resolves the task. Before a fault is injected into a solved trace, the unmodified trace is replayed once and must still resolve the task. The supervisor that proposes subsets, diagnoses, and repairs reads the task, the recorded trajectory, and the failing tests, but never the reference patch.

\subsection{Minimization and repair}
\begin{figure}[t]
\centering
\includegraphics[width=\textwidth]{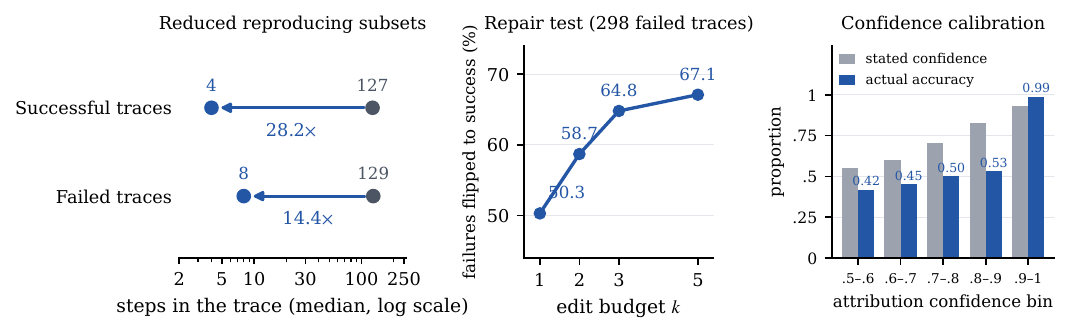}
\par\vspace{-6pt}
\caption{\textbf{The replay study: minimization, repair, and calibration.} Left: median steps of the original trace against its reduced reproducing subset, confirmed by replay against the task's own tests (299 successful and 298 failed traces); the compression factor is the median of the per-trace ratios, not the ratio of the two medians. Center: fraction of the 298 failed traces that flip to success when the supervisor may apply up to $k$ minimal edits at the diagnosed steps. Right: attribution confidence against actual accuracy on the injected-fault corpus (889 scored diagnoses); expected calibration error 0.162.}
\label{fig:minrepair}
\end{figure}

Trace minimization searches for a reduced subset of steps that still reproduces the original outcome under replay (Figure~\ref{fig:minrepair}, left). The median failed trace reproduces its exact failure, with the same failing tests, using 8 of its 129 recorded steps (median per-trace compression $14.4\times$; 78.9\% of 298 traces accept a reduced subset), and the median successful trace still solves the task with 4 of its 127 recorded steps (median per-trace compression $28.2\times$; 87.6\% of 299); most subsets are accepted at the supervisor's first proposal (68.8\% and 80.3\%).

\begin{figure}[t]
\centering
\includegraphics[width=\textwidth]{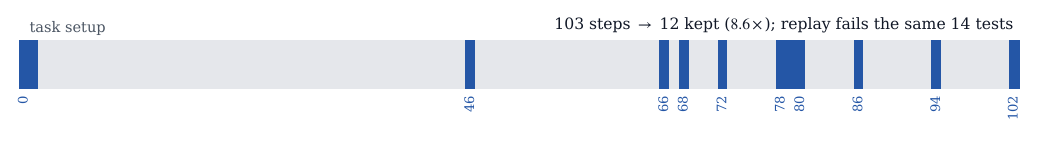}
\par\vspace{-6pt}
\caption{\textbf{A worked minimization example.} Each cell is one step of a 103-step failed trace from the main corpus of the replay study; the 12 blue steps are the reduced subset whose replay fails exactly the same 14 tests as the full trace.}
\label{fig:minexample}
\end{figure}

\textbf{A worked example.} Figure~\ref{fig:minexample} shows one failed trace from the main corpus. The full trajectory has 103 steps and fails 14 of the task's tests. The supervisor's first candidate subset kept 8 steps and a second kept 12, but both replays reproduced only 5 of the 14 failing tests and were rejected: a reduced trace must reproduce the \emph{same} failure, not just some failure. The accepted subset also keeps 12 steps, including ten mid-trajectory actions of which six are writes, and its replay fails exactly the same 14 tests, an $8.6\times$ compression.

In the repair test (Figure~\ref{fig:minrepair}, center), a single-step fix at the diagnosed step flips 50.3\% of 298 failures to success, and allowing up to five edits raises this to 67.1\%, with only 1.65 edits applied on average.

\subsection{The perturbation spectrum}
\begin{figure}[t]
\centering
\includegraphics[width=\textwidth]{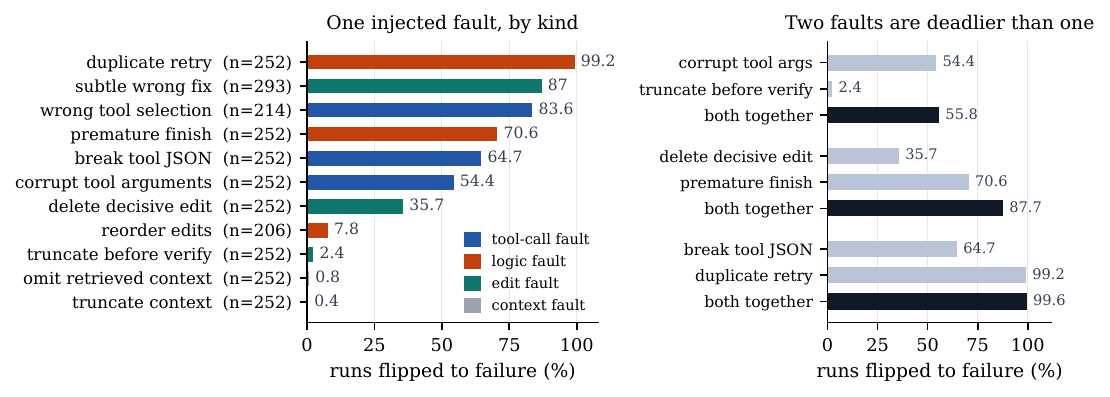}
\par\vspace{-6pt}
\caption{\textbf{The perturbation spectrum.} Left: for eleven of the fifteen fault kinds, the fraction of solved runs that flip to failure when a single fault of that kind is injected and the trace is replayed (3{,}868 valid injections of all fifteen kinds into 297 solved traces), colored by the part of the harness the fault touches. Right: three fault pairs, each fault alone against both injected together (780 paired injections): two faults flip more runs than either alone.}
\label{fig:perturb}
\end{figure}

We injected fifteen kinds of single-step faults into 297 solved traces and replayed each perturbed trace, for 3{,}868 valid injections (Figure~\ref{fig:perturb}). Overall, 35.7\% flip the run to failure, and the spectrum is sharply structured: faults in tool calls, logic, and edits flip 44 to 50\% of runs, faults in context flip 6.7\%, and individual operators range from 99.2\% (duplicating a retry loop) down to 0.4\% (truncating context). Two faults injected at once flip 81.0\% of runs, more than either fault alone. Step-level interventions measure a real, structured quantity, which the \texttt{perturbation} tool exploits.

\subsection{Attribution against ground truth}
Table~\ref{tab:attrgold} scores attribution on 1{,}354 traces where we injected a known fault, so the ground truth holds by construction: the supervisor lists the correct fault class among its candidates 98.7\% of the time, and its first choice is correct 62.0\% of the time, but it names the broken step exactly only 36.9\% of the time and within one step 57.3\% of the time. These 1{,}354 traces are the injections of Figure~\ref{fig:perturb} that flipped a solved run to failure and received a diagnosis (1{,}354 of 1{,}382); the calibration panel of Figure~\ref{fig:minrepair} uses a separate scoring run of 1{,}001 diagnoses, of which the 889 with a single injected fault enter the plot.

On the 208 human-annotated failed traces of CodeTraceBench, drawn from four different agent frameworks, its top step choice lands on a human-flagged step 22.6\% of the time and within three steps of one 39.9\% of the time; a uniformly random step choice would land on a flagged step 9.7\% of the time. Reading identifies what kind of failure occurred far more reliably than where, which is why \method{} gives the optimizer execution tools to test diagnoses and draft edits before submission; whether and how it uses them is part of the optimizer's policy. Attribution confidence is well ordered but overconfident in the middle bins (Figure~\ref{fig:minrepair}, right): the expected calibration error is 0.162, and above confidence 0.85 accuracy reaches 0.95 at 45\% coverage.

\begin{table}[t]
\caption{\textbf{Attribution against constructed ground truth} (1{,}354 traces, each with one injected fault). \emph{Fault class found}: the harness layer of the injected fault is among the layers the supervisor lists (2.9 on average; its first choice is correct in 62.0\% of traces). \emph{Exact step} / \emph{$\pm 1$ step}: the predicted step matches the injected one exactly or within one; a step is one message of the trace. Rows list the six most frequent faults; the overall row covers all injections. When a fault removes content (premature finish, deleted edit), the supervisor names the message right after the injection point (in all 169 and in 75 of 90 traces), which only the $\pm 1$ column counts; a duplicated retry spreads the diagnosis over the retry loop.}
\label{tab:attrgold}
\vspace{7pt}
\centering
\small
\setlength{\tabcolsep}{6pt}
\renewcommand{\arraystretch}{1.08}
\begin{tabular}{l c ccc}
\toprule
Injected fault & $n$ & Fault class found & Exact step & $\pm 1$ step \\
\midrule
corrupt tool arguments & 135 & 100\% & 95.6\% & 95.6\% \\
premature finish & 169 & 100\% & 0\% & 100\% \\
subtle wrong fix & 252 & 100\% & 73.8\% & 76.2\% \\
break tool JSON & 162 & 98.8\% & 50.0\% & 50.6\% \\
delete decisive edit & 90 & 96.7\% & 4.4\% & 87.8\% \\
duplicate retry & 244 & 99.2\% & 0\% & 3.7\% \\
\midrule
overall & 1{,}354 & 98.7\% & 36.9\% & 57.3\% \\
\bottomrule
\end{tabular}
\end{table}

\section{Additional results for Observation 1}
\label{app:obs1}

\subsection{Observation 1 on Terminal-Bench}
\label{app:observations-tb}

Table~\ref{tab:observations-tb} repeats the comparison of Table~\ref{tab:observations} on Terminal-Bench with the small model as the executor and the mid-size model as the optimizer (Appendix~\ref{app:models}), over three evolution rounds. The combination of self-evolution and simple verification is again the strongest configuration in both views, and simple verification alone improves on the frozen optimizer. Self-evolution alone gives no gain: its validation-selected harness scores the same as Meta-Harness ($37.08\%$), and its final harness is within one standard error of it.

\begin{table}[t]
\caption{\textbf{Observation 1 on Terminal-Bench with the small and mid-size models} (avg@3 accuracy, \%, $\pm$SEM over three test evaluations; one evolution run per row). \emph{Self-evolving}: the optimizer edits its own harness; \emph{Verified}: it can run a draft edit on a training task before submitting it. $r^{\star}$: the validation-selected round; \emph{Val-selected}: the harness of that round; \emph{Last}: the final-round harness.}
\label{tab:observations-tb}
\vspace{7pt}
\centering
\small
\setlength{\tabcolsep}{4pt}
\renewcommand{\arraystretch}{1.08}
\begin{tabular}{l cc c cc}
\toprule
Optimizer & Self-evolving & Verified & $r^{\star}$ & Val-selected & Last \\
\midrule
Meta-Harness~\citep{metaharness2026} & \no & \no & 3 & $37.08_{\pm1.94}$ & $37.08_{\pm1.94}$ \\
Self-evolving & \yes & \no & 2 & $37.08_{\pm1.29}$ & $37.83_{\pm1.98}$ \\
Simple verification only & \no & \yes & 2 & $38.95_{\pm0.75}$ & $39.33_{\pm1.12}$ \\
\textbf{Self-evolving $+$ simple verification} & \yes & \yes & 2 & $\mathbf{39.33}_{\pm0.65}$ & $\mathbf{40.45}_{\pm0.65}$ \\
\bottomrule
\end{tabular}
\end{table}

\subsection{Self-evolution with and without verification, round by round}
\label{app:case}

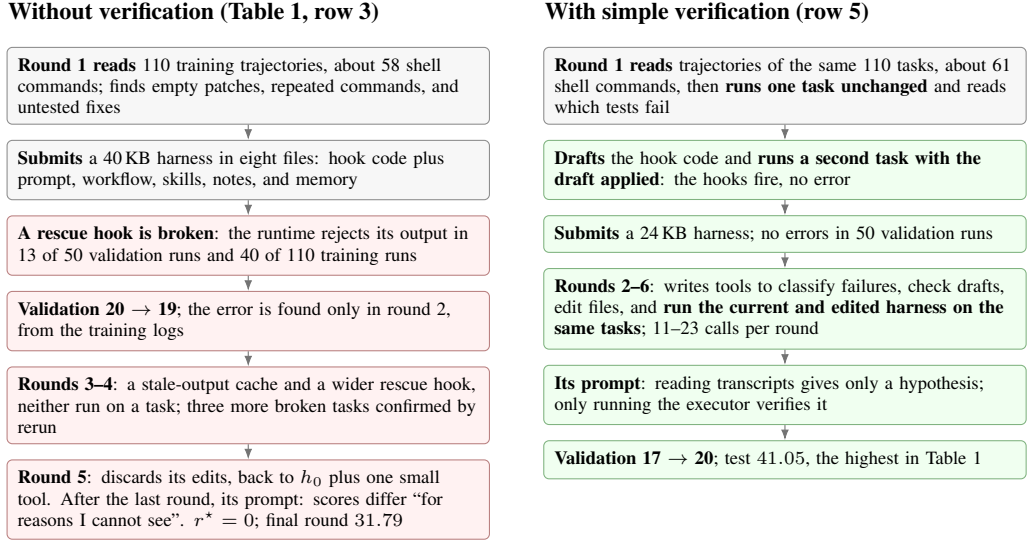
\begin{figure}[t]
\centering
\begin{tikzpicture}[font=\scriptsize,
  sb/.style={draw=black!45, rounded corners=2pt, align=left, text width=6.1cm, inner sep=4pt, fill=black!3},
  bad/.style={sb, fill=red!5, draw=red!45!black!50},
  good/.style={sb, fill=green!6, draw=green!40!black!50},
  lane/.style={font=\small\bfseries, anchor=south west},
  arr/.style={-{Latex[length=1.4mm]}, black!55}]
\node[lane] at (-0.1,0.2) {Without verification (Table~\ref{tab:observations}, row 3)};
\node[sb] (a1) at (3.2,-0.5) {\textbf{Round 1 reads} 110 training trajectories, about 58 shell commands; finds empty patches, repeated commands, and untested fixes};
\node[sb, below=2mm of a1] (a2) {\textbf{Submits} a 40\,KB harness in eight files: hook code plus prompt, workflow, skills, notes, and memory};
\node[bad, below=2mm of a2] (a3) {\textbf{A rescue hook is broken}: the runtime rejects its output in 13 of 50 validation runs and 40 of 110 training runs};
\node[bad, below=2mm of a3] (a4) {\textbf{Validation 20 $\to$ 19}; the error is found only in round 2, from the training logs};
\node[bad, below=2mm of a4] (a5) {\textbf{Rounds 3--4}: a stale-output cache and a wider rescue hook, neither run on a task; three more broken tasks confirmed by rerun};
\node[bad, below=2mm of a5] (a6) {\textbf{Round 5}: discards its edits, back to $h_0$ plus one small tool. After the last round, its prompt: scores differ ``for reasons I cannot see''. $r^{\star}=0$; final round $31.79$};
\foreach \i/\j in {a1/a2,a2/a3,a3/a4,a4/a5,a5/a6} \draw[arr] (\i) -- (\j);
\node[lane] at (7.0,0.2) {With simple verification (row 5)};
\node[sb] (b1) at (10.3,-0.5) {\textbf{Round 1 reads} trajectories of the same 110 tasks, about 61 shell commands, then \textbf{runs one task unchanged} and reads which tests fail};
\node[good, below=2mm of b1] (b2) {\textbf{Drafts} the hook code and \textbf{runs a second task with the draft applied}: the hooks fire, no error};
\node[good, below=2mm of b2] (b3) {\textbf{Submits} a 24\,KB harness; no errors in 50 validation runs};
\node[good, below=2mm of b3] (b4) {\textbf{Rounds 2--6}: writes tools to classify failures, check drafts, edit files, and \textbf{run the current and edited harness on the same tasks}; 11--23 calls per round};
\node[good, below=2mm of b4] (b5) {\textbf{Its prompt}: reading transcripts gives only a hypothesis; only running the executor verifies it};
\node[good, below=2mm of b5] (b6) {\textbf{Validation 17 $\to$ 20}; test $41.05$, the highest in Table~\ref{tab:observations}};
\foreach \i/\j in {b1/b2,b2/b3,b3/b4,b4/b5,b5/b6} \draw[arr] (\i) -- (\j);
\end{tikzpicture}
\par\vspace{-6pt}
\caption{\textbf{The two self-evolving runs of Table~\ref{tab:observations}, without and with simple verification}, reconstructed from the optimizer's tool calls and round logs. Red: edits submitted without being run on a task, and their consequences; green: steps based on running the executor.}
\label{fig:casefull}
\end{figure}

Figure~\ref{fig:casefull} follows the two self-evolving runs of Table~\ref{tab:observations} round by round, reconstructed from the optimizer's tool calls, the round logs, and the prompts, notes, and tools the optimizer wrote for itself. The two runs use the same host, models, number of candidates, turn limit, and training tasks, and both read the trajectories and scores of the training tasks. The only difference is that one run has simple verification, with a budget of eight calls per candidate (Appendix~\ref{app:models}).

\textbf{Without the tool.} In round 1, the three candidates each spent about 58 shell commands reading the 110 training trajectories. The winner found three failure causes: empty patches, repeated commands, and fixes that the repository's own tests never checked. It submitted a 40\,KB harness in eight files: hook code, a system prompt, a workflow, three skills, tool notes, and a memory file.

One hook, meant to rescue runs that stop early, returned output that the hook runtime rejects. The error appeared in 13 of the 50 validation runs and in 40 of the 110 training runs of the next round. Validation accuracy fell from 20 to 19 of 50 tasks, and a rerun confirmed one broken task. The optimizer found the error only in round 2, from the training logs. In rounds 3 and 4, it added a cache that answered repeated commands with stale output and then widened the rescue hook; it ran neither on a task before submitting it, and reruns confirmed three more broken tasks.

Before round 4, it wrote its own tool for checking drafts, which loads the hook code and feeds it made-up model replies. This tool could show that the code ran, but not whether the executor did better on real tasks. From round 3 on, its notes told it to ``stop paying for code layers''. In round 5, it discarded its edits and returned to the initial harness plus one small file-editing tool; in round 6, it added one hook. After the last round, it wrote in its own prompt that harnesses differing by 9 to 30\,KB of code ``score 3 apart for reasons I cannot see''. Validation accuracy never rose above its round-0 value, so validation selected the initial harness itself ($r^{\star}=0$ in Table~\ref{tab:observations}). The final-round harness scored $31.79\%$.

\textbf{With the tool.} In round 1, the candidates read the trajectories of the same 110 training tasks with about 61 shell commands each. The winner then ran one training task under the unchanged harness and read which tests failed. It drafted its hook code and ran another task with the draft: the hooks fired without error. The submitted 24\,KB harness raised no errors in the 50 validation runs.

In later rounds, the optimizer wrote its own tools: one that classifies failed training runs, one that checks draft code, one that edits files, and a probe that runs the current and the edited harness on the same training tasks. The three candidates called these tools 11 to 23 times per round. The optimizer's prompt came to state that reading transcripts gives only a hypothesis, and that only running the executor verifies it. Validation accuracy rose from 17 to 20 of 50 tasks, and the final harness scored $41.05\%$, the highest in Table~\ref{tab:observations}.

\textbf{The difference.} Both optimizers read trajectories of the same training tasks and found similar failure causes in round 1. What differed was whether they could test an edit before submitting it. Without the tool, errors surfaced only after a round had been spent, and the optimizer could not tell which of its edits changed validation accuracy. Its self-evolution turned toward doing less, until it discarded its edits. With the tool, the optimizer checked its drafts on real tasks, and its self-evolution went into better ways of testing them.

\section{Additional results for Observation 2}
\label{app:artifacts}

The audit covers the five runs of Figure~\ref{fig:census}, one per executor: Qwen3.8-27B (the last row of Table~\ref{tab:observations}), Qwen3-Coder-480B, DeepSeek V3.2, GLM-5, and Mistral Large 3. Each run evolves six rounds with the self-evolving optimizer and simple verification. The optimizer workspace has five editable slots: a prompt file added to the optimizer's own instructions, a candidate-guidance file whose blocks are handed to its parallel candidates, a notes file, a skills directory, and one Python file that may declare callable tools and automatic hooks on the optimizer's own agent loop. All five runs start with the three files blank, the skills directory empty, and no Python file. We read the workspace after the final round.

An artifact is one self-written unit: a prompt file, a skill file, a callable tool declared in the Python file, an automatic hook, or the notes file. An automatic hook is one check or reminder with its own trigger condition; a series of reminders keyed only to the turn counter counts once. Each artifact receives one primary function: attribution (locating and ranking failure causes), verification (checking a draft before submission), training audit (recording outcomes across rounds), or workflow (how the optimizer organizes its own process, including budget, turn allocation, and the division of work among its candidates). Superseded versions, temporary scripts, executor harness files, and the pipeline's own reports are not counted. Table~\ref{tab:artifact-inventory} lists all 51 artifacts.

Three of the five runs cover all four functions; the GLM-5 run wrote no attribution artifact and the Mistral Large 3 run no verification artifact. Every run kept a training audit and workflow artifacts, but the implementations differ. Two runs (DeepSeek V3.2 and GLM-5) wrote no callable tool and placed their verification logic in submission-time hooks instead; the Qwen3.8-27B run wrote the largest toolkit, including a probe tool that batches head-versus-edit comparisons over simple verification calls. Every run rewrote both prompt files and kept a notes file, and the notes are where confirmed fixes and regressions, prediction outcomes, and process lessons accumulate.

Presence does not establish correctness or benefit. The Qwen3.8-27B run's own notes record a triage tool that raised on every call in one round and a probe tool that misread transcripts until a later revision, and the Python file is loaded only if it passes the same safety check as the executor's hooks and a test run, so a rejected version leaves the previous one in place.

\begin{figure}[t]
\centering
\includegraphics[width=0.6\linewidth]{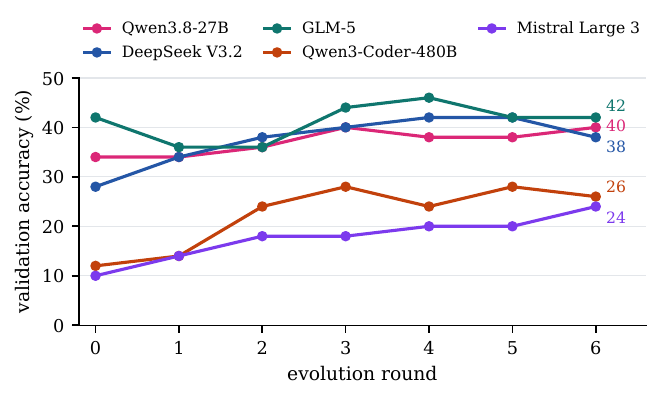}
\par\vspace{-12pt}
\caption{\textbf{Validation accuracy per evolution round} of the self-evolving optimizer with simple verification, run from scratch on SWE-rebench with the five executors of Figure~\ref{fig:census}. Validation pool: 50 tasks per round; round 0 is the initial harness.}
\label{fig:valcurves}
\end{figure}

Figure~\ref{fig:valcurves} plots the validation accuracy of these five runs across evolution rounds. The best validation round exceeds round 0 on every executor, and the final round stays above round 0 on four of the five.

{\renewcommand{\arraystretch}{0.9}\renewcommand{\footnotesize}{\fontsize{8.45}{9.5}\selectfont}\LTnormalcaption\setlength{\linewidth}{\dimexpr(\textwidth-24pt)*100/89\relax}\footnotesize
\setlength{\tabcolsep}{3pt}
\begin{longtable}{@{}>{\raggedright\arraybackslash}p{0.13\linewidth}>{\raggedright\arraybackslash}p{0.075\linewidth}>{\raggedright\arraybackslash}p{0.215\linewidth}>{\raggedright\arraybackslash}p{0.105\linewidth}>{\raggedright\arraybackslash}p{0.365\linewidth}@{}}
\caption{\textbf{Inventory of self-written optimizer artifacts} behind Figure~\ref{fig:census}. One row is one artifact; the identifier is the file in the optimizer workspace, the tool name declared in \texttt{extra\_tools}, or the hook point that implements an automatic check. Each artifact carries one primary function. Descriptions summarize the retained content and do not assert that it worked.}
\label{tab:artifact-inventory}\\
\toprule
Executor & Type & Identifier & Function & Content \\
\midrule
\endfirsthead
\toprule
Executor & Type & Identifier & Function & Content \\
\midrule
\endhead
Qwen3-Coder-480B & Prompt & prompt.md & Workflow & Eleven standing rules: read notes first, triage in bulk, validate before submitting, keep edits additive. \\
 & Prompt & search.md & Workflow & Three focus blocks that assign a different target to each parallel candidate. \\
 & Skill & skills/\allowbreak{}triage\_\allowbreak{}playbook.md & Attribution & Count failure signatures before opening transcripts; cross-check notes for categories already fixed. \\
 & Skill & skills/\allowbreak{}validate\_\allowbreak{}and\_\allowbreak{}ship.md & Verification & Run 2--3 failing and 1--2 passing tasks with the draft before writing the submission. \\
 & Skill & skills/\allowbreak{}hooks\_\allowbreak{}audit\_\allowbreak{}constraints.md & Verification & The safety-check rules a hook file must pass; lint before submitting. \\
 & Skill & skills/\allowbreak{}destructive\_\allowbreak{}mechanism\_\allowbreak{}caution.md & Verification & Higher validation bar for mechanisms that mutate or revert the executor's files. \\
 & Tool & \texttt{triage\_\allowbreak{}scan} & Attribution & Counts failure-signature substrings across a trajectory directory in one call. \\
 & Tool & \texttt{lint\_\allowbreak{}hooks} & Verification & Regex check of a draft hook file against the executor's safety-check rules. \\
 & Hook & on\_\allowbreak{}turn\_\allowbreak{}end & Workflow & Turn 8: read notes and triage; turn 20: spend probes; turn 30: lint before submitting. \\
 & Notes & notes.md & Training audit & Per-round outcomes, ranked failure categories, process lessons, and an open task queue. \\
\midrule
DeepSeek V3.2 & Prompt & prompt.md & Workflow & Structural facts about the loop, the current baseline state, and standing rules for each round. \\
 & Prompt & search.md & Workflow & Three focus blocks: repair last round's regression, audit the full hook stack, target turn exhaustion. \\
 & Skill & skills/\allowbreak{}investigation.md & Attribution & Grep recipes for failure classes; separates root causes from symptoms; checks cross-round oscillation. \\
 & Skill & skills/\allowbreak{}validation.md & Verification & Probe protocol: failing targets, passing regression tasks, one unrelated repository, then calibrate predictions. \\
 & Hook & before\_\allowbreak{}tool(submit) & Verification & Returns the submission once if predicted fixes were never probed with the final edits. \\
 & Hook & before\_\allowbreak{}tool(submit) & Verification & Returns the submission once if probes covered fewer than three repositories. \\
 & Hook & on\_\allowbreak{}turn\_\allowbreak{}end & Workflow & Reminds after six probe calls that the pool is shared across candidates. \\
 & Notes & notes.md & Training audit & Per-round fixed and regressed tasks, harness changes, and two structural findings about selection. \\
\midrule
GLM-5 & Prompt & prompt.md & Workflow & Test a new regex-based hook on synthetic strings before spending probes. \\
 & Prompt & search.md & Workflow & Safe-fallback, diversity, and alternate-mechanism slots for the three candidates. \\
 & Skill & skills/\allowbreak{}regression\_\allowbreak{}safety.md & Verification & Risk tiers for edits, false-positive probes that must reach the guarded branch, prediction rules. \\
 & Hook & before\_\allowbreak{}tool(submit) & Verification & Flags benchmark task identifiers inside submitted file content. \\
 & Hook & before\_\allowbreak{}tool(submit) & Verification & Requires probes with edits on at least two distinct tasks before the first submission attempt. \\
 & Hook & before\_\allowbreak{}tool(submit) & Verification & Flags predicted fixes that the optimizer's own probes with these edits returned as failing. \\
 & Hook & before\_\allowbreak{}tool(submit) & Verification & Asks for evidence that a probe reached a new blocking hook before it ships. \\
 & Hook & before\_\allowbreak{}tool(submit) & Verification & Suggests shipping a revert alone rather than bundled with a new mechanism. \\
 & Hook & on\_\allowbreak{}turn\_\allowbreak{}end & Workflow & Reserves remaining probes for passing-task regression checks after six calls. \\
 & Notes & notes.md & Training audit & Post-mortems of rounds 5 and 6: unexercised branches, entangled reverts, screening ties. \\
\midrule
Mistral Large 3 & Prompt & prompt.md & Workflow & Eight standing instructions: read memory, triage in bulk, budget probes, prefer narrow edits. \\
 & Prompt & search.md & Workflow & Candidate A revives a validated but unshipped mechanism; B finds a new signature; C is free. \\
 & Skill & skills/\allowbreak{}triage.md & Attribution & Find the highest-leverage mechanical cause in the first turns; weight aggregate evidence. \\
 & Tool & \texttt{survey\_\allowbreak{}failures} & Attribution & Batch-scans trajectory files for a list of regex signatures and reports counts with examples. \\
 & Hook & before\_\allowbreak{}tool & Workflow & Parses a stringified edits argument into a list, or blocks the call with an explanation. \\
 & Hook & on\_\allowbreak{}turn\_\allowbreak{}end & Workflow & Every four probe calls: commit to one confirmed hypothesis and reserve regression checks. \\
 & Hook & on\_\allowbreak{}turn\_\allowbreak{}end & Workflow & After 25 shell calls: switch to the batch scanning tool. \\
 & Notes & notes.md & Training audit & Per-round root causes and outcomes, process lessons, and a checklist for the next round. \\
\midrule
Qwen3.8-27B & Prompt & prompt.md & Workflow & What is scored, how selection behaves, evidence rules, edit-design rules, and a turn schedule. \\
 & Prompt & search.md & Workflow & A shared floor for all candidates plus three focus blocks: shrink the current harness, two failure groups. \\
 & Skill & skills/\allowbreak{}corpus-mine.md & Attribution & Two-call corpus digest: bucket runs by verdict and audit which current-harness checks fire on passes. \\
 & Skill & skills/\allowbreak{}loss-mode-map.md & Attribution & Maps each failure bucket to the harness levers that can and cannot move it. \\
 & Skill & skills/\allowbreak{}verifier-verdicts.md & Attribution & Reads the grader's verdict taxonomy; tests that never ran and tests that failed need opposite fixes. \\
 & Skill & skills/\allowbreak{}evidence-protocol.md & Verification & A/B before proposing, measure task noise with repeats, keep a passing regression task, falsify the edit. \\
 & Skill & skills/\allowbreak{}harness-anatomy.md & Workflow & Verified facts about the executor loop, hook surface, knobs, audit rules, and scoring. \\
 & Tool & \texttt{triage} & Attribution & Failure map by grader verdict with cheap probe targets; grep and bucket views on request. \\
 & Tool & \texttt{probe} & Verification & Runs the current and the edited harness on the same tasks with repeats and reports per-run telemetry. \\
 & Tool & \texttt{lint} & Verification & Runs the pipeline's real audit function over a draft hook file. \\
 & Tool & \texttt{edit} & Workflow & Anchored patching of a draft with all-or-nothing anchors; stages the file for probing. \\
 & Hook & on\_\allowbreak{}turn\_\allowbreak{}end & Workflow & Scheduled notes at turns 3, 8, 15, 21, 26, and 30+: triage, probe, draft, A/B, assemble, submit. \\
 & Hook & before\_\allowbreak{}tool /\allowbreak{} after\_\allowbreak{}tool & Workflow & Answers a third identical shell command from cache instead of rerunning it. \\
 & Hook & on\_\allowbreak{}turn\_\allowbreak{}end & Workflow & Reports failures of the self-written tools so their silence is not mistaken for evidence. \\
 & Notes & notes.md & Training audit & Round outcomes with reproduced flips, selection behaviour, the harness's own failure record. \\
\bottomrule
\end{longtable}
\normalsize
}

\section{Examples of optimizer-written artifacts}
\label{app:examples}

This appendix shows what the optimizer wrote for itself in our best run: \method{} with self-evolution on the Meta-Harness host, with Qwen3.8-Flash-Next as the optimizer and Qwen3.8-27B as the executor (Table~\ref{tab:equip-open}; $42.28\%$ in distribution and $37.69\%$ out of distribution). Validation selected round 2. We show the optimizer harness that produced this round, which the optimizer wrote in its self-update after round 1. Its notes, skills, and code were empty at the start; \method{} seeds only its prompt and candidate guidance with instructions for the verification tools, and Appendix~\ref{app:ex-skill} shows only the lines the optimizer added to them. Excerpts are verbatim, except that internal names are replaced by the terms used in this paper; omissions are marked \texttt{[...]}.

The logs of this run show three lessons behind its result. First, the optimizer measured before it edited. In round 1, it classified the 79 failed training runs: 41 ran tests with a Python interpreter that lacked the test framework, 28 tried to install packages offline, and 26 tried to reach the network. Its round-1 edit to the executor harness addressed these failure classes as a whole rather than individual tasks. When the new harness ran on the training tasks, the three counts fell to 2, 0, and 0, and training accuracy rose from 31 to 37 of 110 tasks. In its self-update, the optimizer turned this analysis into a tool it could call in one step (Appendix~\ref{app:ex-tool}).

Second, the optimizer learned from a regression it caused. The round-1 edit fixed three validation tasks but broke two that had passed. In its self-update, the optimizer traced this to its verification calls, which had tested only tasks it expected the edit to fix. It added a rule to its prompt, skill, and notes: check every draft on two tasks it should fix and one task the current harness already passes (Appendices~\ref{app:ex-skill} and~\ref{app:ex-notes}). It also wrote a hook that blocks, once, a submission made without executed evidence (Appendix~\ref{app:ex-hooks}).

In round 2, all three verification calls included a passing task, compared with none of the 3 calls in round 1 and none of the 12 calls in the run with the same host and tools but without self-evolution. The round-2 edit fixed both earlier regressions and four other validation tasks and broke one, raising validation accuracy from 17 to 22 of 50 tasks. It also added exceptions to the round-1 interpreter rewrite for projects that manage their own environments, a change the notes had listed as a lead.

Third, once a failure class was gone, the optimizer moved to the next one. After round 1, the largest remaining class was runs whose visible tests passed but whose hidden tests failed (30 of 73 failures). The notes list this as the next lead (Appendix~\ref{app:ex-notes}), and the round-2 edit made the executor check its patch against the requirements stated in the issue and run the wider test directory before submitting.

\subsection{A tool for measuring failures}
\label{app:ex-tool}
The optimizer wrote \texttt{corpus\_triage} in its first self-update, after spending 56 to 62 shell calls per candidate on this analysis in round 1. The tool classifies all failed training runs in one call and lists tasks that the current harness passes, which the optimizer then uses as regression checks.

\begin{codeblock}{Tool declaration and entry point (\texttt{hooks.py}, excerpt)}
def extra_tools(self):
    return [{
        "name": "corpus_triage",
        "description": ("Run the standard failure-mode triage over the latest round's "
                        "trajectories in ONE call: pass/fail counts, failure-class flags "
                        "(never_ran_tests, no_edit_seen, base_python_no_pytest, "
                        "pip_try_offline, green_but_graded_fail, mangled_nodeids, "
                        "net_attempt) with example task ids, and the passing-task list to "
                        "pick regression canaries from. Replaces dozens of exploratory "
                        "bash calls. Optional arg: dir (path to a trajectories dir)."),
        "input_schema": {"type": "object",
                         "properties": {"dir": {"type": "string",
                                                "description": "trajectories dir (default: latest round)"}}},
    }]

def run_tool(self, name, args, env, state):
    if name != "corpus_triage":
        return "unknown tool %s" % name
    try:
        d = ((args or {}).get("dir") or "").strip().replace("'", "")
        argv = " - %s" % d if d else " -"
        return env.bash("python3%s <<'TRIAGEPY'\n%s\nTRIAGEPY\n" % (argv, _TRIAGE))
    except Exception as e:
        return "corpus_triage failed: %r (fall back to manual bash triage)" % (e,)
\end{codeblock}
\begin{codeblock}{Failure-class flags computed by the tool (excerpt)}
ran = any(re.search(r"pytest|-m unittest", c) for c in cmds)
green = bool(re.search(r"\d+ passed", t)) and not re.search(r"\d+ failed", t)
base = bool(re.search(r"No module named .?pytest", t))
pip = any(re.search(r"\b(pip|uv|conda)\b.*install", c) for c in cmds)
edit = any(re.search(r"sed -i|write_text|git apply|patch -p|\.write\(", c) for c in cmds)
rows.append((tid, ok))
if not ok:
    for k, v in [("never_ran_tests", not ran), ("no_edit_seen", not edit),
                 ("base_python_no_pytest", base), ("pip_try_offline", pip),
                 ("green_but_graded_fail", green), ("mangled_nodeids", "ERROR: not found" in t),
                 ("net_attempt", "NewConnectionError" in t)]:
        if v: flag[k].append(tid)
\end{codeblock}

Running the same script on the training runs that enter round 2 gives the following output.

\begin{textblock}{Output on the training runs that enter round 2}
triage over trajectories: 110 tasks, 37 pass, 73 fail
green_but_graded_fail     30  e.g. ['CQCL__guppylang-1259', 'Spenhouet__confluence-markdown-exporter-92', [...]]
mangled_nodeids           15  e.g. ['All-Hands-AI__OpenHands-10628', 'davep__braindrop-96', [...]]
no_edit_seen               4  e.g. ['avaiga__taipy-2797', 'bvanelli__actualpy-117', [...]]
never_ran_tests            3  e.g. ['avaiga__taipy-2797', 'bvanelli__actualpy-117', [...]]
base_python_no_pytest      2  e.g. ['Spenhouet__confluence-markdown-exporter-92', 'sissbruecker__linkding-1291', [...]]
passing set (37): pick regression canaries from these
ASPP__pelita-863, DiamondLightSource__mx-bluesky-1519, PlasmaControl__DESC-1901, PyPSA__linopy-438, [...]
\end{textblock}

The optimizer also summarized what went wrong in round 1 at the top of its code. Its count of about 110 shell calls per session is its own: round 1 was interrupted and restarted, and the log it read held both attempts, each with 53 to 62 shell calls per candidate.

\begin{textblock}{Header of \texttt{hooks.py}}
"""Governor hooks for MY optimization loop (authored after round 1).

Measured failure patterns of my own optimization sessions (round 1 evidence):
  * ~110 bash corpus-mining calls per candidate session vs 4-6 verification calls: the
    scarce resources (turns, shared execution pool) were spent on exploration
    that a single reusable triage script does in one call.
  * The accepted change-set's transforming hooks fixed 3 tasks but REGRESSED 2
    previously-passing ones: probes only tested predicted-fix directions, never
    a "stays green" canary from the currently-passing set.
  * 4 of 9 observed flips did not reproduce: single-run flip evidence is ~coin
    flip; conclusions should weight confirmed (twice-run) flips.
  * Every session still submitted (good) - keep the anti-deadlock releases.

Design: every hook is fail-open (identity on any exception), blocks are capped
and self-releasing, and the only additive tool runs a pure-computation triage
script over the run directory.
"""
\end{textblock}

\subsection{Hooks on the optimizer's own loop}
\label{app:ex-hooks}
Before each tool call, one hook caps the verification calls of each candidate session so that parallel sessions share the pool, caps exploratory shell calls, and blocks a submission once if no draft has been verified yet. Two further hooks inject reminders at fixed turns and append rules to the optimizer's system prompt, each citing the round-1 measurement behind it.

\begin{codeblock}{Budget and submission gates (\texttt{before\_tool}, excerpt)}
BASH_SOFT = 40          # bash mining calls allowed before guidance kicks in
BASH_HARD = 70          # absolute cap once drafting/probing is underway
POOL_CAP = 4            # verification + simple_verification calls per session (pool shared across sessions)

def before_tool(self, name, args, state):
    try:
        if name in ("verification", "simple_verification"):
            used = state.get("pool", 0)
            if used >= POOL_CAP:
                return ("execution pool reserved",
                        "BLOCKED: this session already used its %d shared-execution slots "
                        "(other parallel sessions need them). Revise your draft using "
                        "evidence you already have, then submit." % POOL_CAP)
            state["pool"] = used + 1
            return args

        if name == "bash":
            [...]   # caps corpus-mining bash calls (40 soft / 70 hard), anti-deadlock release

        if name == "submit_proposal":
            edits = []
            try:
                edits = (args or {}).get("edits") or []
            except Exception:
                return args
            if edits and not state.get("pool") and not state.get("warned"):
                state["warned"] = True
                return ("probe the draft first",
                        "BLOCKED (once): zero executed evidence yet this session. Draft "
                        "these same bytes in /scratch and verify 2 predicted-fix targets "
                        "+ 1 passing canary; a probe reporting your hook call counts beats "
                        "any amount of reading. Then resubmit (this block fires only once).")
            return args
    [...]
\end{codeblock}
\begin{codeblock}{Turn milestones (\texttt{on\_turn\_end}, excerpt)}
def on_turn_end(self, turn, state):
    [...]
    try:
        if t == 20 and not state.get("draft"):
            return ("Milestone t20: stop broad mining. Write your top-2 root causes and "
                    "draft edit files to /scratch now (corpus_triage covers the stats).")
        if t == 32 and not state.get("pool"):
            return ("Milestone t32: no executed evidence yet. Verify your current draft "
                    "on 2 fix-targets + 1 canary NOW; pool left this session: %d."
                    % max(0, POOL_CAP - state.get("pool", 0)))
        if t == 45:
            return ("Milestone t45: final probe (same bytes as submission) then SUBMIT "
                    "within a few turns. Unsubmitted = 0 for the whole session.")
        if t == 52:
            return ("SUBMIT NOW with your best current change-set - budget is almost out. "
                    "A mediocre submitted change-set beats a perfect unsubmitted one.")
\end{codeblock}
\begin{codeblock}{Rules appended to the system prompt (\texttt{system\_prompt})}
def system_prompt(self, assembled):
    try:
        return (assembled or "") + (
            "\n\nBUDGET DISCIPLINE (measured round 1: ~110 mining bash calls per session "
            "vs 4 probes - backwards). (1) Use the corpus_triage tool (or the one triage "
            "script) FIRST; then targeted greps only. Commit to your top 2 root causes by "
            "turn ~25 and draft edits in /scratch. (2) Every probe batch must include at "
            "least one REGRESSION CANARY: a task the CURRENT harness already passes - "
            "round-1 transforming hooks fixed 3 and regressed 2 because no canary was "
            "ever probed. (3) A single-run flip is ~50% noise: count only twice-confirmed "
            "flips; prefer killing a whole failure class over chasing task luck. "
            "(4) Reserve the last ~8 turns: final probe of the exact bytes you submit, "
            "then submit. Unsubmitted session = zero.")
\end{codeblock}

\subsection{A skill and prompt rules}
\label{app:ex-skill}
The optimizer wrote one skill in its first self-update, on writing executor hooks safely. Rule~5 is the regression check it derived from round 1: test each draft on failing tasks and on one task that already passes. It also added the following lines to its seeded prompt.

\begin{textblock}{Skill: \texttt{skills/hook\_safety.md}}
# Skill: writing executor hooks that don't blow up the run

The executor's harness_code/hooks.py runs on EVERY run of EVERY task. One bad
transform is a run-wide regression; one raise per call is reported in your audit.

## Contract facts (verify in round_NN/executor_source/{executor.py,hooks_runtime.py})
- A run ends when a reply has no tool_use block and stop_reason=end_turn; the score
  is the accumulated `git diff`. after_llm can append a tool_use block to keep the
  run alive -- append ONLY `text`/`tool_use` typed blocks (a `thinking` block in
  the returned list made the API validator reject the reply in a real incident).
- before_tool returning a (reason, synthetic_result) tuple blocks the command and
  injects the synthetic observation: zero container cost, zero wasted turn.
- after_tool sees the raw observation before the model; the model's own commands are
  visible in before_tool args. State dict persists per run.

## Safety rules
1. Fail open: wrap every method body in try/except returning the input unchanged;
   a hook that raises degrades to identity AND prints an error in your report card.
2. Act on tight, anchored patterns, never broad keywords; bail out (return input)
   the moment anything unexpected appears (quoted heredocs, pipes, env vars,
   explicit alternative interpreters like `uv`/`poetry`/venv paths -- an
   interpreter-rewrite that ignores these sank previously-passing tasks once).
3. Prefer annotating an observation to rewriting a command; prefer blocking a
   never-can-work action (offline fetch) to transforming a maybe-valid one.
4. One-shot messages: gate the gate. A submit-gate must fire once, not loop the
   run; a keep-alive must count its rescues (cap 2-3).
5. Prove it with verification on 2 failing targets + 1 passing canary. Read the probe's
   per-hook call/changed counts and tracebacks -- called 112x/changed 0x means your
   condition never matched (dead code), not success.
6. Perturb before you codify: only distill a behavior into a hook/skill when
   perturbation shows it is load-bearing in passing trajectories.
\end{textblock}
\begin{textblock}{Lines added to the optimizer's prompt after round 1 (excerpt)}
0. Your notes.md already summarizes the current harness and last round's measured failure classes -- start there, then run the `corpus_triage` tool (one call: pass/fail counts, failure-class flags, passing-task list for canaries). Do targeted greps only after that. Commit to your top-2 root causes by turn ~25 and start drafting edit files in /scratch; a full re-derivation of the stats wastes the half of the session where probes live.

1. [...] BEFORE submitting, draft your edits and verify them on 2 tasks you predict they fix PLUS 1 REGRESSION CANARY -- a task the current harness already PASSES. Historical predicted-fix hit rates without testing are 0-16%; and a round-1 proposal that probed only fix-directions fixed 3 tasks and REGRESSED 2 previously-passing ones. [...]

2. Flip evidence is noisy: ~half of single-run flips fail to reproduce on rerun (r1: 5/9). One twice-confirmed flip beats three single-run flips; prefer mechanisms that kill a WHOLE failure class (they survive the noise) over cleverness aimed at one task.

## Session pacing (self-enforced; the governor hook will nag you)
- t<=20: triage + targeted greps; write root-cause hypotheses + intended edits to /scratch.
- t 20-35: draft all edit files; probe #1 (2 fix targets + 1 canary).
- t 35-45: revise from probe output (read the hook call counts / tracebacks, not just PASS/FAIL); probe #2 with final bytes.
- t 45+: submit byte-identical to the probed bytes; if nothing is ready, submit your best unprobed change-set anyway (better than zero).
- Shared pool: at most 4 verification/simple_verification calls per session -- parallel sessions and the finalist session need the rest.
\end{textblock}

\subsection{Notes}
\label{app:ex-notes}
After round 1, the notes record the optimizer's scores, its confirmed fixes and regressions, the failure classes it measured on the training runs, the rules it derived, and leads for the next round. Two sections, which describe the current harness and the executor's loop, are omitted. The notes are the optimizer's own account, and two of their statements differ from the logs. The training accuracy listed for the round-1 harness, 31 of 110, is that of the initial harness; the round-1 harness solved 37 of 110 when it ran in round 2. The notes also mention a second session for the winning candidate; the protocol runs none, and the only other session of a round is the self-update.

\begin{textblock}{Notes: \texttt{notes.md} (excerpt)}
# Optimizer memory (cross-round) -- condensed, keep updated every self-update

## Scoreboard
- r0 harness: val 16/50, train 31/110.
- r1 (kept, MIXED): env-normalising + patch-protecting hooks.py + prompt set -> val 17/50, train 31/110. Net +3 fixed / -2 regressed; only 5/9 flips reproduced on rerun (confirmed fixed: 2 of the claimed; the litellm-style REGRESSION confirmed real). LESSON: single-run flips ~coin flips; transforming hooks carry real regression risk.
- r1 screening mechanics: 3 candidate sessions with the 3 candidate-guidance lenses; each ~55-60 turns; winner gets a second (final) session; execution pool (verification + simple_verification) shared round-wide (~24 slots; 11 used). Winner r1 = lens 1 (OMISSION).

[...]

## Corpus facts (r1 triage, 110 tasks, use corpus_triage tool for fresh numbers)
- 31 pass / 79 fail. Flags: base_python_no_pytest ~41-50, pip_try_offline 28, net_attempt 26, green_but_graded_fail 18, never_ran_tests 17, mangled_nodeids 16, no_edit_seen 14.
- Classes: FAILED-with-a-patch ~40 (CORRECTNESS gap -- biggest untouched, next frontier), NOT_RUN ~22 (several UNWINNABLE: the verifier's own pytest dies -- offline uv, git-apply conflict, parametrized id truncated in eval script; don't chase), NO_PATCH 13, REGRESSION 3.
- 16 failures had fully green last pytest run -> graded by hidden tests matching issue wording (names/messages/edges). "Existing tests pass" is not the bar.
- Passing runs (mean ~29 turns): grep-locate -> sed/python edit -> repo-env pytest on touched file -> whole dir -> git diff -> submit. 0/31 skipped tests; 17/79 fails never ran tests.

[...]

## Method rules (hard-won -- obey)
1. Verify the DRAFT BYTES before submit: 2 predicted-fix targets + >=1 REGRESSION CANARY (task the current harness passes). Submit byte-identical to the probed set. r1's biggest miss: never probed a canary -> -2 regressions.
2. A refuted probe downgrades specific task claims, not the mechanism (empty-diff -> real-patch = mechanism verified even at 0 flips).
3. Max 2 probes per hypothesis, then move on; never burn the shared pool (>4/session starves the final session).
4. Never end without submit_proposal. An unsubmitted session = 0.
5. ~50% of single-run flips are noise; the report card reruns flips -- build claims that survive twice.
6. [...] minimize/perturb on FAILING traces rarely flips -- perturb PASSING traces to find load-bearing behaviors and distill them into skills.
7. Don't raise obs_cap/max_turns for free -- check context pressure first (r1 left loop mostly default).

## Open leads for next round
- CORRECTNESS (40 FAILED-with-patch): help the executor hold the diff against the issue's assertions -- e.g. an extra tool that diffs requirements vs patch (metered self-review via env.llm), or a run_tests tool that also prints the touched module's full test results compactly. Verify with the verification tool (correctness flips are noisy -- probe twice if pool allows).
- Regression-proofing the current harness: guard transforms (skip rewrite when the command mentions an explicit different interpreter/venv/uv/poetry; skip gate if a prior submit was blocked already).
- NO_PATCH survivors: keep-alive exists; check whether the rescue commands actually run (after_llm changed 0/112 calls in probes -- rescue FIRES but the appended command may not be executed? investigate before adding more).
- Verifier-NOT_RUN tasks are dead budget: identify from verifier output and blacklist mentally (they rotate; re-identify each round from fresh output, not from stale ids).
\end{textblock}

\end{document}